\documentclass[runningheads]{llncs}
\usepackage{graphicx}

\usepackage{bbding}
\usepackage{tikz}
\usepackage{comment}
\usepackage{amsmath,amssymb} 
\usepackage{color}
\usepackage{url}
\usepackage{epsfig}
\usepackage{subfigure}
\usepackage{multirow}
\newcommand{\MaleLight}[0]{Male Light}
\newcommand{\MaleDark}[0]{Male Dark}
\newcommand{\FemaleLight}[0]{Female Light}
\newcommand{\FemaleDark}[0]{Female Dark}

\makeatletter
\newcommand{\printfnsymbol}[1]{%
  \textsuperscript{\@fnsymbol{#1}}%
}
\makeatother

\begin{document}
\pagestyle{headings}
\mainmatter
\def\ECCVSubNumber{100}  

\title{FairFace Challenge at ECCV 2020: \\Analyzing Bias in Face Recognition} 

%
\author{Tom\'aš Sixta~\inst{1,\thanks{These (corresponding~\Envelope) authors contributed equally to this work.}} \and
Julio C. S. Jacques Junior\inst{2,3,\printfnsymbol{1}} \and
Pau Buch-Cardona\inst{3,4} \and Neil M. Robertson\inst{5} \and 
Eduard Vazquez\inst{6} \and
Sergio Escalera\inst{3,4}}
\authorrunning{T. Sixta et al.}
%
\institute{Czech Technical University in Prague, Czech Republic -- \email{tomas.sixta@gmail.com}
\and Universitat Oberta de Catalunya, Spain --
\email{jsilveira@uoc.edu}
\and Computer Vision Center, Spain 
\and Universitat de Barcelona, Spain -- \email{sergio@maia.ub.es}
\and The Queen's University of Belfast, United Kingdom -- \email{N.Robertson@qub.ac.uk}
\and Anyvision, United Kingdom -- \email{eduardov@anyvision.co}
}


\maketitle
\setcounter{footnote}{0}

\begin{abstract}
This work summarizes the 2020 ChaLearn Looking at People Fair Face Recognition and Analysis Challenge and provides a
description of the top-winning solutions and analysis of the results. The aim of the challenge was to evaluate accuracy and bias in gender and skin colour of submitted algorithms on the task of 1:1 face verification in the presence of other confounding attributes. Participants were evaluated using an in-the-wild dataset based on reannotated IJB-C, further enriched by 12.5K new images and additional labels. 
The dataset is not balanced, which simulates a real world scenario where AI-based models supposed to present fair outcomes are trained and evaluated on imbalanced data. The challenge attracted 151 participants, who made more than 1.8K submissions in total. The final phase of the challenge attracted 36 active teams out of which 10 exceeded 0.999 AUC-ROC while achieving very low scores in the proposed bias metrics. Common strategies by the participants were face pre-processing, homogenization of data distributions, the use of bias aware loss functions and ensemble models. The 
analysis of 
top-10 teams shows higher false positive rates (and lower false negative rates) for females with dark skin tone as well as the potential of eyeglasses and young age to increase the false positive rates too.
\end{abstract}


\section{Introduction}

Automatic face recognition is a general topic that includes both face identification and verification~\cite{JAYARAMAN2020}. Face identification is the process of identifying someone's identity given a face image. This process is generally known as 1-to-n matching and could be seen as asking to the system ``who is this person?''. Face verification, on the other hand, is concerned with validating a claimed identity based on the image of a face, and either accepting or rejecting the identity claim (1-to-1 matching). A simple example of face verification is when people unlock their smartphones using their faces (e.g., authentication), whereas searching for the identity of a given individual in a database of missing people, for instance, could be an example of face identification. 

Fairness in face recognition recently started to receive increasing interest from different segments of scientific communities \cite{Drozdowski2020,LoPiano2020,Mehrabi2019,Pierce2020}. This is partially due to the huge impact new technologies have in our daily lives. Face recognition has been routinely utilized by both private and governmental organizations around the world \cite{Davies2018,ValentinoDeVries2020}. Automatic face recognition can be used for legitimate and beneficial purposes (e.g. to improve security) but at the same time its power and ubiquity heightens a potential negative impact unfair methods can have for the society \cite{Uyghurs2018,Uyghurs2019,Mozur2020,Raji2020}. Recently, these concerns led several major companies to suspend distribution of their products to US police departments until a legislation regulating its deployment is passed by US Congress~\cite{CongressAmazon2020,CongressMicrosoft2020,CongressIBM2020}. 

Although not sufficient, a necessary condition for a legitimate deployment of face recognition algorithms is equal accuracy for all demographic groups. A gold standard for testing commercial products is the Face Recognition Vendor Test (FRVT) performed by National Institute of Standards and Technology (NIST) \cite{Grother2019,Snow2018}. However, this test is not designed for iterative and fast evaluation of new research directions. There is also a growing number of works that evaluate the algorithms on public data \cite{Albiero2020c,WangMei2018,Srinivas2019,Cavazos2019} and are therefore limited by what data is available, i.e., typically either small scale high quality datasets or large scale datasets with noisy annotations.

To motivate research on fair face recognition and provide a new challenging accurately annotated dataset, we designed and ran a computational face recognition challenge where participants were asked to provide solutions that maximize both accuracy and two fairness scores (minimize bias score). The submissions were evaluated on a reannotated version of IJB-C \cite{Maze2018} database, enriched by newly collected 12,549 public domain images. The dataset contains large variations in head pose, face size and other attributes (detailed in Sec.~\ref{sec:database}). 
The dataset is not balanced with respect to different attributes, which imposes another challenge for the participants and is intended to stimulate usage of bias mitigation methods, also because the final ranking is defined by a weighted combination of accuracy and fairness (giving the bias scores a higher weight). To this end, we propose a new evaluation metric derived from a causal model by means of a causal effect of protected attributes to the accuracy of the algorithm, detailed in Sec.~\ref{sec:evaluationmethod}. The challenge attracted a total of 151 participants, who made more than 1.8K submissions in total\footnote{Data and winning solutions codes are available at \urlstyle{rm}\url{http://chalearnlap.cvc.uab.es/challenge/38/description}}. 
We expect the provided dataset and proposed fairness measure template to be a reference evaluation benchmark for face recognition systems, and that the outcomes of this challenge will help both to define priorities for future research as well as to help on the definition of technical requirements for real applications.



\section{Ethics in Face Recognition}\label{sec:ethics}

Face recognition methods have been researched for decades due to their wide number of scenarios for good\footnote{For more information about ethics in AI you can visit the European guideline in the following link \urlstyle{rm}\url{https://ec.europa.eu/digital-single-market/en/news/ethics-guidelines-trustworthy-ai}.}. They can be applied, e.g., in robotics, human-computer interaction, access and control, security, among others. Recently, face recognition research received additional attention due to the improved performance provided by deep learning architectures~\cite{GUO2019102805}. When it comes to public safety, 
past works raised the question about the efficacy of facial recognition systems for law enforcement following the apparent failure of the systems to identify suspects, reporting as possible reasons for failure problems like occlusions, angled facial shots, poor lighting or obscured facial features by hats or sunglasses~\cite{20131}. However, recent studies show that automated methods for face analysis can also discriminate based on classes like gender and ethnicity~\cite{Buolamwini2018}, among others, which raised an additional focus of attention around such technologies. If face recognition methods are used to support decisions, erroneous but confident mis-identification can have serious consequences, and these possible and negative outcomes are making the society to rethink about what should be the limits of such technology, especially when it is applied at larger scales involving additional privacy concerns.

From a research point of view, a bottleneck to be solved is to develop methods that can work accurately for all target populations. 
While there is a need to promote good practices and reinforce regulations, we need to find the way to provide the required good (and fair) performance in practice,
and if face recognition is to be applied, it should deal with the bias problem. Evidences show that the computer vision and machine learning research communities are starting to give visibility to different types of bias~\cite{Bird:2019:FML,friedler2019comparative} and proposing different solutions to mitigate them (e.g.,~\cite{Torralba:cvpr:2011,Buolamwini2018,Escalante:TAC:2020,Hendricks_2018_ECCV,Wang_2019_ICCV,Yucer_2020_CVPR_Workshops}). 
Nonetheless, additional efforts should be made to further reduce bias in future methods. This is precisely the main goal of the 2020 ChaLearn Looking at People Fair Face Recognition and Analysis Challenge, i.e., to stimulate and promote research on face recognition methods that produce fair outcomes.

\section{Related Work}\label{sec:relatedwork}
It is known that popular face recognition datasets like Labeled Faces in the Wild (LFW) \cite{LearnedMiller2016}, MegaFace \cite{Kemelmacher2016}, IJB-C \cite{Maze2018}, IMDB-WIKI \cite{Rothe2015,Rothe2018}, VGGFace2 \cite{Cao2017} or MS-Celeb-1M \cite{Guo2016} are imbalanced both in gender and skin colour \cite{Merler2019}. To encourage research in fair face recognition there is growing number of datasets specifically designed with balance in mind and annotated for gender, ethnicity and potentially other attributes. Examples are Racial Faces in the Wild (RFW) \cite{WangMei2018} (40K images, 12K identities, subset of MS-Celeb-1M), Balanced Faces in the Wild (BFW) \cite{Robinson2020} (20K images, 0.8K identities, subjects sampled from VGGFace2) or DiveFace \cite{Morales2019} (150K images, 24K identities, subset of Megaface). Even though these datasets are important step towards fairer face recognition, using labels for ethnicity does not in general allow for comparing models across datasets, because unlike for skin colour \cite{Bino2013} there is no widely accepted definition of ethnicity groups and the labels instead rely on judgment of the annotators. Furthermore, balancing alone may not be enough to guarantee fair models \cite{Albiero2020b}, which motivates research of bias mitigation methods.

Nowadays, the gold standard for evaluating accuracy and bias of face recognition algorithms is the ongoing FRVT Test performed by NIST~\cite{Grother2019,Snow2018}. 
The submitted (mostly commercial) algorithms are evaluated on four datasets composed of photographs from various visa/benefits US governmental applications. In total, there are 18.27 million images of 8.49 million people. Besides FRVT, there are numerous small scale evaluations of bias in publicly and commercially available algorithms (e.g. \cite{WangMei2018,Srinivas2019,Cavazos2019}) as well as analysis of bias in models trained from scratch on publicly available datasets \cite{Albiero2020c}, that in most cases report better accuracy for men and people with light skin colour. 

Traditional measures of fairness are based on calculating certain statistics related to the error rate of the algorithm. For example, Equalized Odds requires the true positive and false positive rates to be equal for all protected groups (see \cite{Verma2018,Drozdowski2020} for a comprehensive review). These measures are easy to calculate, but without having background in statistics it can be difficult to choose the ``correct'' one for the task at hand. This is a serious shortcoming, because certain traditional measures in general contradict each other \cite{Berk2017,Chouldechova2016,Kleinberg2016} as it was dramatically demonstrated on the case of COMPAS (a system used in some US states to predict the risk of recidivism) \cite{Angwin2016,Compas2016}. Individual Fairness \cite{Dwork2011} tries to overcome these shortcomings by deriving a fairness measure from intuition, that ``\textit{similar individuals are treated similarly}''. However, it does not provide any general definition of similarity and only postpones the problem by proposing that it should be given by a regulatory body or a civil rights organization. 

A growing number of state of the art measures is based on causal inference. They require a ``model of the world'' given as a causal diagram and the actual measure is then derived using this diagram, e.g. in terms of the causal effect of the protected attributes to the algorithm accuracy \cite{Zhang2016,Nabi2017} or using counterfactuals \cite{Kusner2017}, i.e., would the decision remain had the value of the protected attribute be different but everything else stayed the same.
A crucial advantage of these approaches is that the underlying ethical views are encoded by the diagram in an easy to understand way, which exposes them to criticism and allows them to be changed if they prove to be inadequate.
Furthermore, as these approaches are trying to identify the true causes of the unfairness, they can be used as a starting point for mitigating the bias in the real world.

Bias mitigation methods can be broadly divided based on what area of model deployment they target to pre-processing, in-processing and post-processing \cite{Bellamy2018,Pessach2020}. The most popular pre-processing technique is rebalancing the dataset \cite{Huang2018,WangZeyu2020}, alternatively using synthetic data \cite{Kortylewski2019}. In-processing approaches include cost-sensitive training (higher weights for underrepresented groups) \cite{Huang2018}, adversarial learning for removing the sensitive information from the features \cite{Alvi2018,WangZeyu2020}, tuning parameters of a loss function for different protected groups \cite{Morales2019,WangMei2020} or attempts to learn bias free representations in unsupervised way \cite{Vowels2020}. Examples of post-processing techniques are renormalizing the similarity score of two feature vectors based on the demographic groups of the corresponding images \cite{Terhorst2020b} or attaching more fully connected layers to the feature extractor in order to remove the sensitive information from the representations \cite{Morales2019}. 
The FairFace Recognition challenge, described in Sec.~\ref{sec:challengedesign}, did not impose any constrain to the participants to what model stage bias mitigation should be addressed. The best solutions rely on a combination of different strategies, detailed in Sec.~\ref{sec:resultswinning}.

\vspace{-0.2cm}
\section{Challenge Design}\label{sec:challengedesign}
\vspace{-0.2cm}

The participants were asked to develop their face verification methods aiming for a reduced bias in terms of the protected attributes (i.e., gender and skin color). Developed methods needed to output a list of confidence scores  given test ID pairs to be verified (higher score means higher confidence, that the image pair contains the same person). 
The challenge\footnote{\urlstyle{rm}\url{https://competitions.codalab.org/competitions/24184}} was managed using Codalab\footnote{\urlstyle{rm}\url{https://competitions.codalab.org}}, an open source framework for running competitions that allows result or code submission. 

The challenge ran from 4th April to 1st July 2020, and included two different phases: development and test. In the development phase, the participants were provided with public train data (with labels) and validation data (without labels, from which they should make predictions). At the test stage, the validation labels were released to all participants as well as the test data (without labels, considered for the final evaluation). 
The challenge attracted a total of 151 registered participants. During development phase we received 1330 submissions from 48 teams, and 476 submission from 36 teams at the test stage, resulting in more than 1800 submissions in total.
Additional schedule details and participation statistics are provided in the supplementary material.




\vspace{-0.2cm}
\subsection{The Dataset}\label{sec:database}

The dataset used in the challenge is a reannotated version of IJB-C \cite{Maze2018}, further enriched by newly collected 12,549 public domain images. In total, there are 152,917 images from 6,139 identities. The images were annotated by Anyvision's internal annotation team for two protected attributes: gender (male, female) and skin colour (light corresponding to Fitzpatrick types I-III, dark corresponding to types IV-VI) and five legitimate attributes: age group (0-34, 35-64, 65+), head pose (frontal, other), image source (still image, video frame), wearing glasses and a bounding box size\footnote{Attribute categories used in this work are imperfect for many reasons. For example, it is unclear how many skin colour and gender categories should be stipulated (or whether they should be treated as discrete categories at all). We base our definitions on widely accepted traditional categories and our methodology and findings are expected to be applied later to any re-defined and/or extended attribute category.}. Detailed annotation instructions are in the supplementary material.
Every attribute 
was annotated by at least 3 annotators (age and skin colour by 6, due to their subjectiveness, aiming to maximize the level of agreement). 
Labels for gender and skin colour were synchronized for each \emph{identity} to the most prevalent ones, and labels of the other attributes were obtained by choosing for each \emph{image} the most common label from the annotators. 

\begin{figure*}[htbp]
  \centering
    \subfigure[Positive pairs]
    {\includegraphics[height=1.4cm]{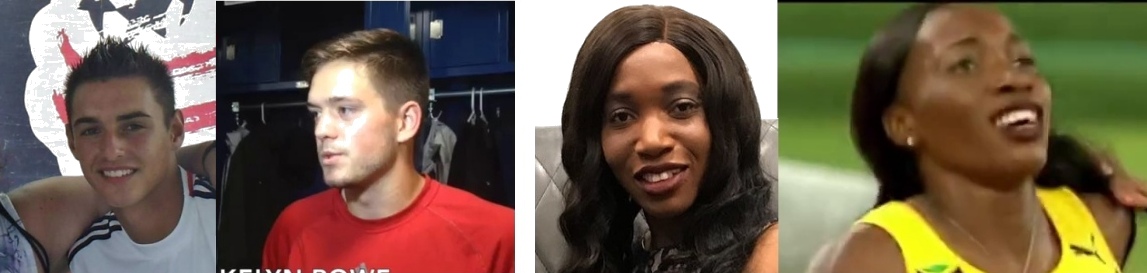}}
    \hspace{0.5cm}
    \subfigure[Negative pairs]
    {\includegraphics[height=1.4cm]{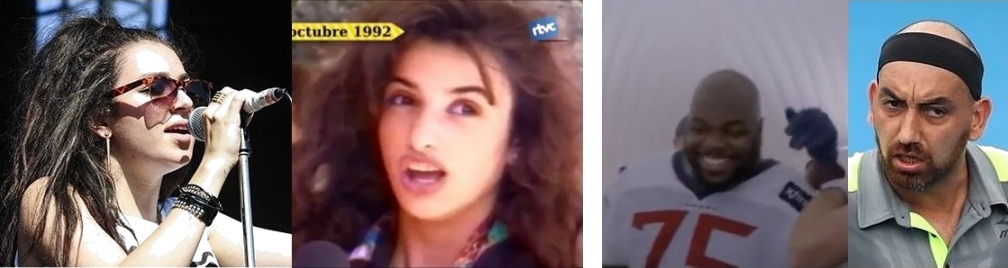}}
    \vspace{-0.4cm}
    \caption{Positive and negative samples of image pairs used in the challenge.}
\label{fig:dataset:examples}
\end{figure*}

\begin{figure*}[htbp]
  \centering
    \subfigure[Gender \& Skin Colour]
    {\includegraphics[height=2.65cm]{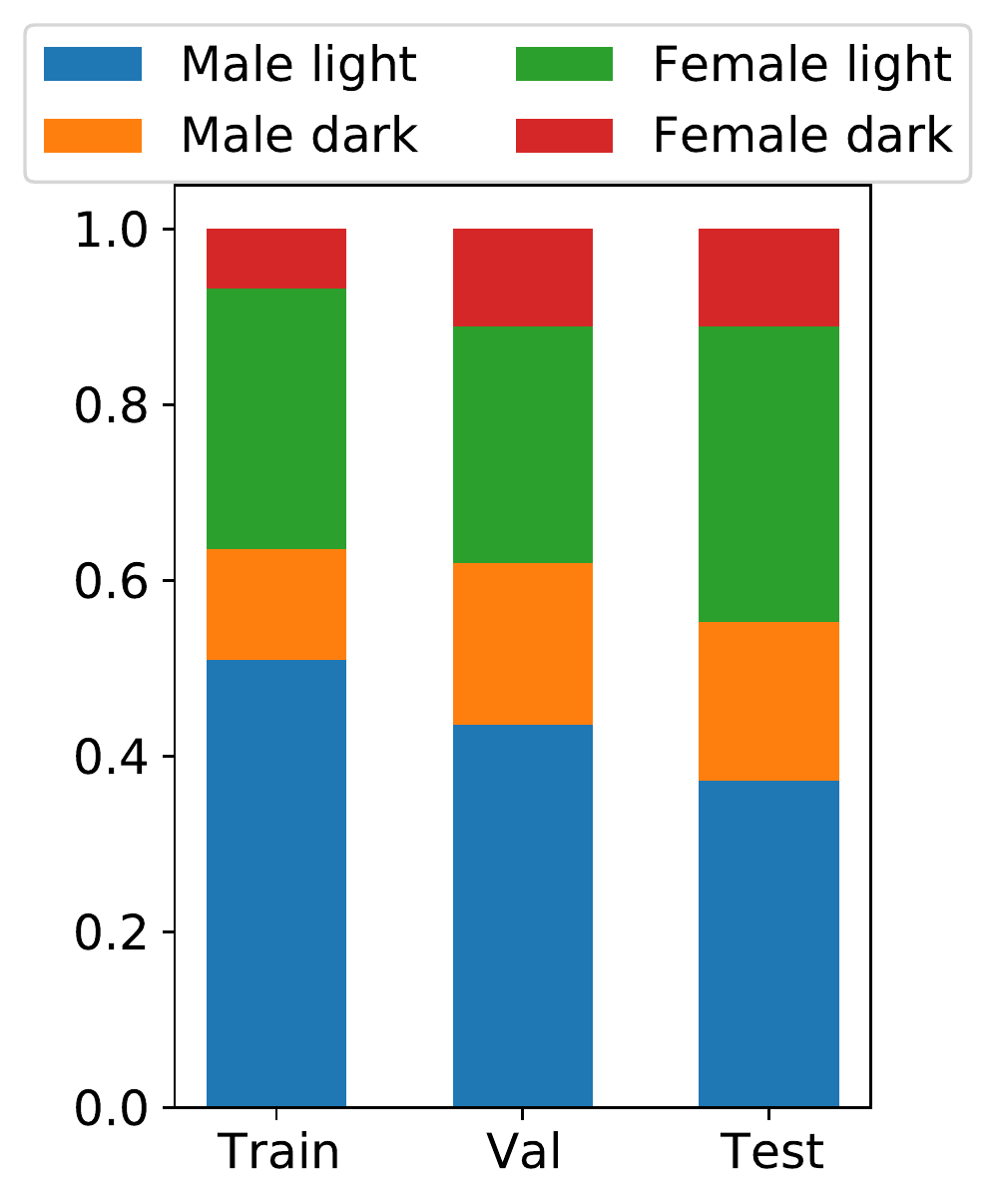}
      \label{fig:dataset:attributes:protected}
    }
    \subfigure[Age]
    {\includegraphics[height=2.65cm]{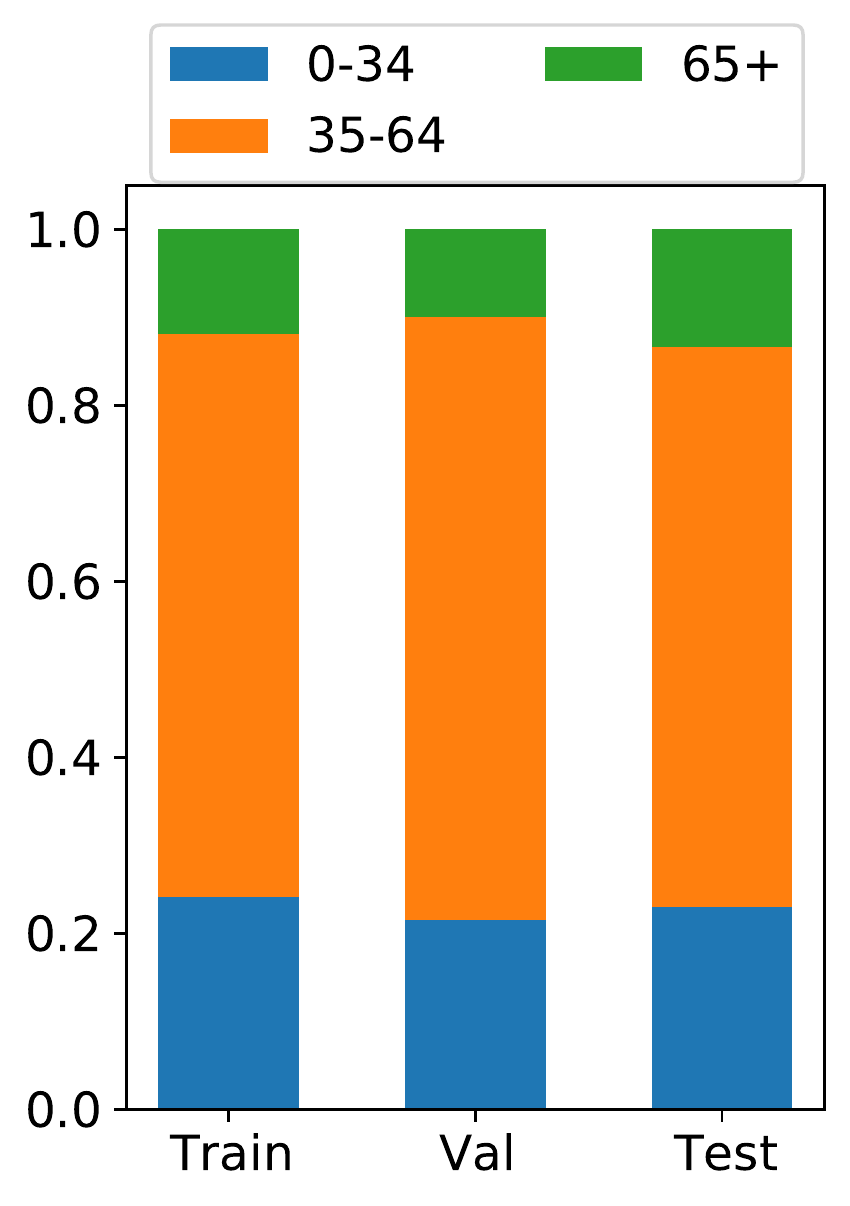}}
    \subfigure[Glasses]
    {\includegraphics[height=2.48cm]{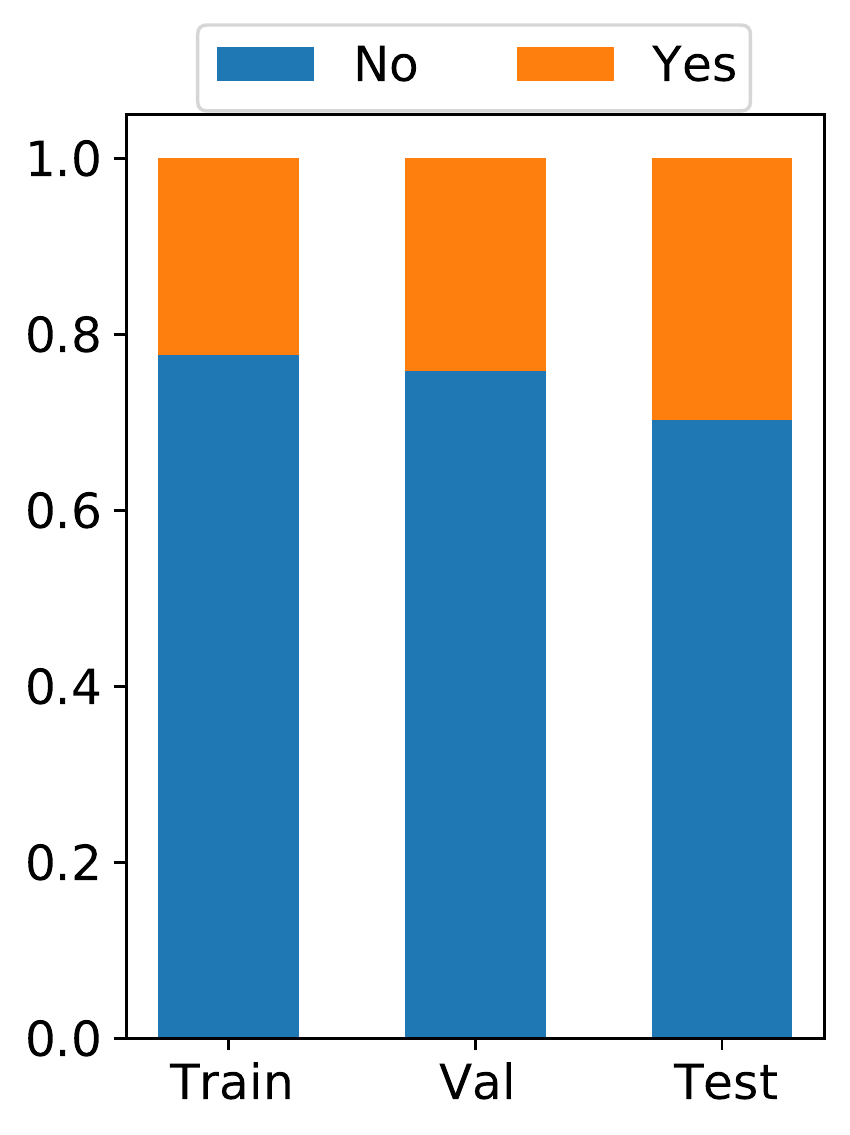}}
    \subfigure[Pose]
    {\includegraphics[height=2.48cm]{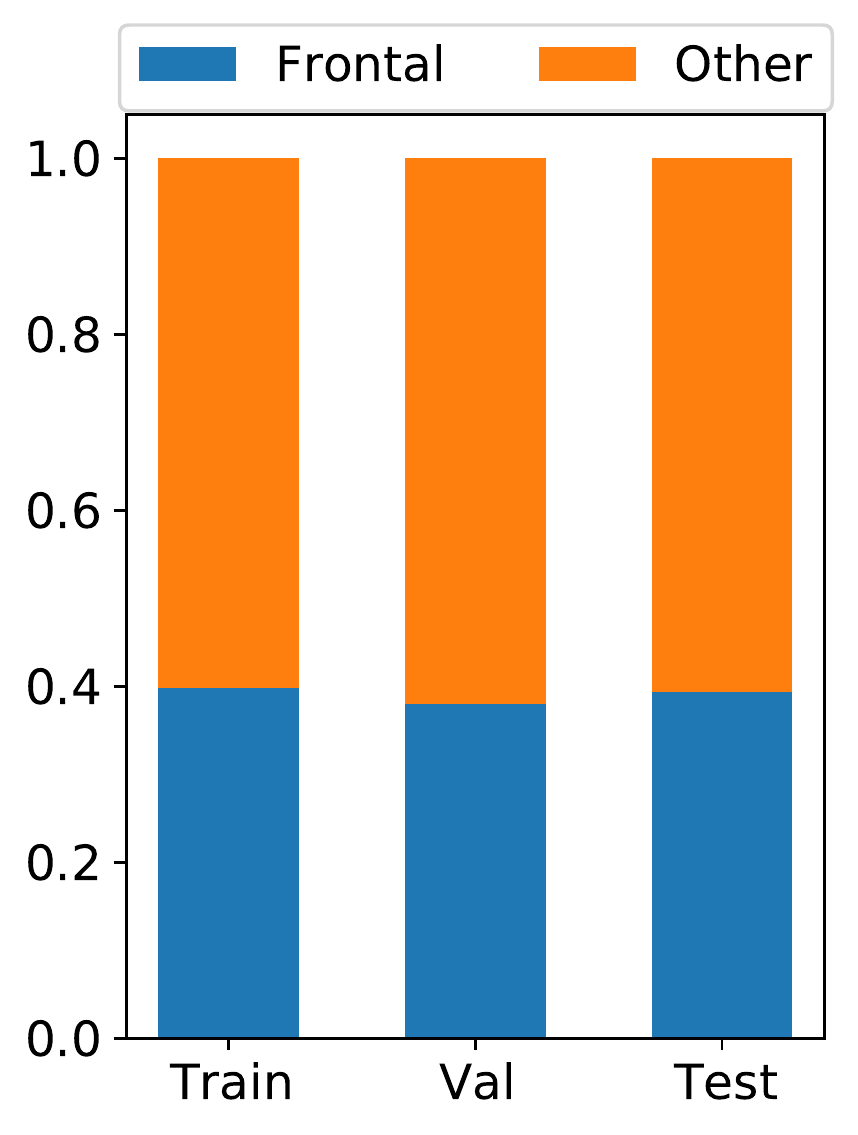}}
    \subfigure[Source]
    {\includegraphics[height=2.48cm]{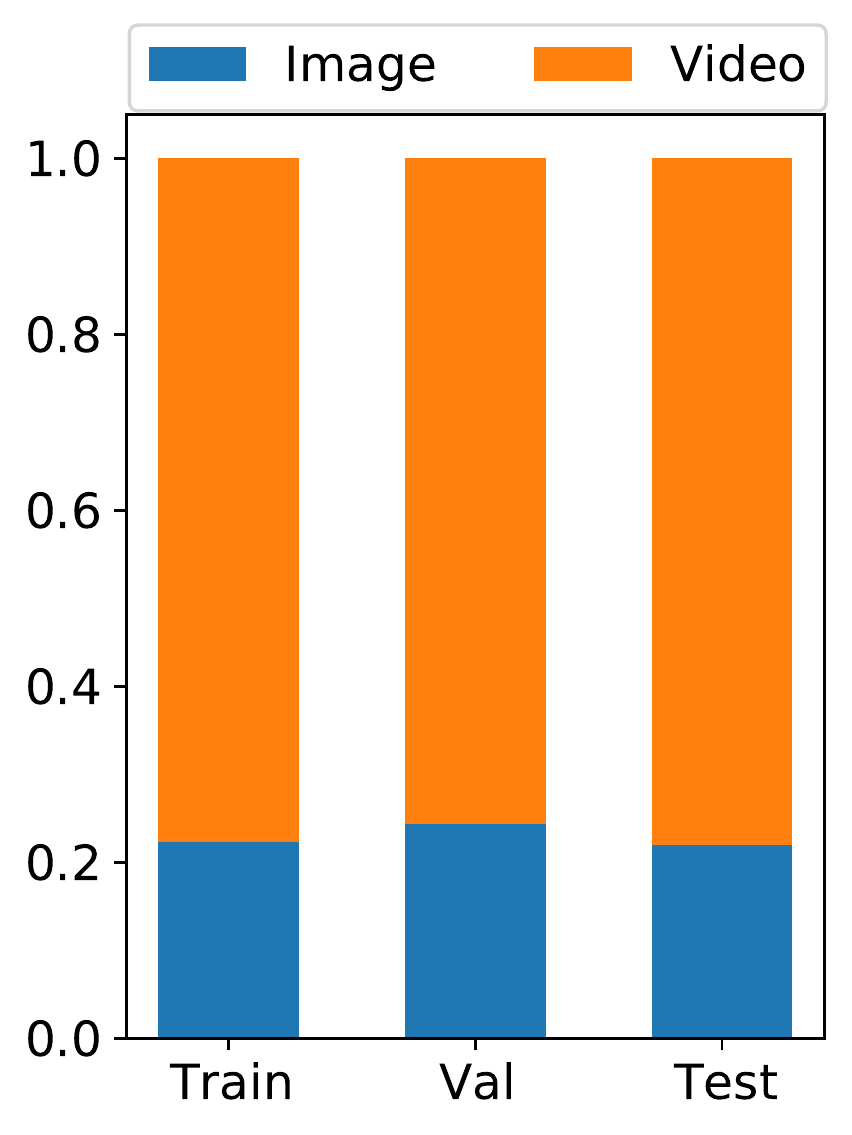}}
    \subfigure[B-Box]
    {\includegraphics[height=2.48cm]{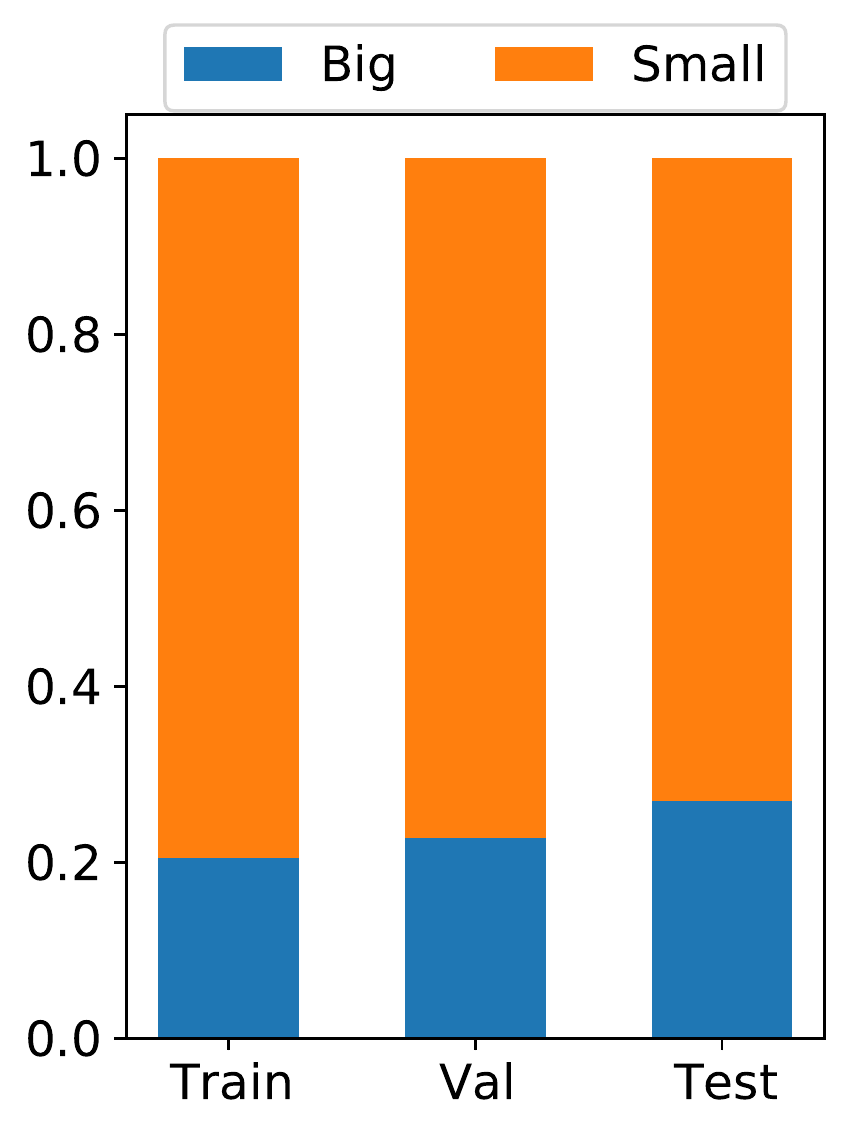}}
    \caption{Distribution in percentage of attributes in training, validation and testing subsets of the dataset. Bounding box of a face was considered small if either its width or height was smaller than 224 px. }
\label{fig:dataset:attributes}
\end{figure*}

For the purpose of the challenge, the dataset was split into training, validation and testing subsets containing 70\%, 10\% and 20\% of identities. To facilitate evaluation of the submitted results we generated roughly half a million positive face image pairs (same identity) and half a million negative pairs for both validation and testing subsets. The pairs were selected such that the number of combinations of legitimate attributes is maximized. In the validation pairs there were 219 (positive) and 574 (negative) combinations, and test pairs contained 397 (positive) and 1162 (negative) combinations. Basic dataset statistics are summarized in Table \ref{t:dataset:statistics}. 
Few image samples and and statistics of the attributes are shown in Fig.~\ref{fig:dataset:examples} and Fig.~\ref{fig:dataset:attributes}, respectively.

\begin{table}[htbp]
	\centering
			\setlength\tabcolsep{4pt}
	\scriptsize
\caption{Dataset statistics.}
\begin{tabular}{|l|c|r|r|c|}
\hline
 & \multicolumn{1}{l|}{\textbf{Train}} & \multicolumn{1}{l|}{\textbf{Validation}} & \multicolumn{1}{l|}{\textbf{Test}} & \multicolumn{1}{l|}{\textbf{Total}} \\ \hline
\textbf{Images} & \multicolumn{1}{r|}{100,186} & 17,138 & 35,593 & \multicolumn{1}{r|}{152,917} \\ \hline
\textbf{Unique identities} & \multicolumn{1}{r|}{4,297} & 614 & 1,228 & \multicolumn{1}{r|}{6,139} \\ \hline
\textbf{Positive pairs} & - & 448,119 & 500,176 & - \\ \hline
\textbf{Negative pairs} & - & 552,672 & 500,963 & - \\ \hline
\end{tabular}
\label{t:dataset:statistics}
\end{table}

The images in the dataset have large variance in head pose, bounding box size and other attributes, which makes it challenging for face recognition. At the same time the distribution of these attributes is imbalanced, for example as seen in Fig.~\ref{fig:dataset:attributes:protected}, there is considerably more white males that dark females. Such imbalances are common in real world datasets and we intentionally have not reballanced the data to encourage research of bias mitigation methods.

\vspace{-0.5cm} 
\subsection{Evaluation Protocol}\label{sec:evaluationmethod}
\begin{figure}[htbp]
	\centering
	\includegraphics[height=2.2cm]{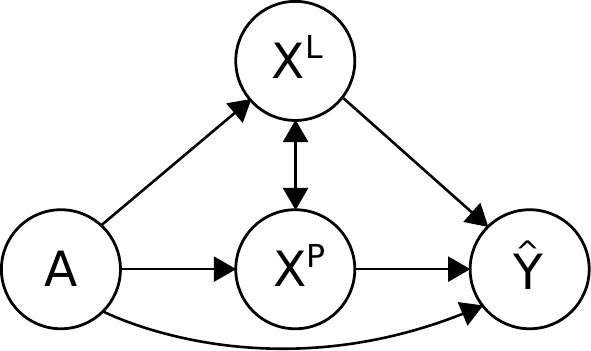}
    \caption{Causal model used for our definition of fairness. $A$: protected attributes (gender and skin colour), $X^L$: legitimate attributes, $X^P$: proxy attributes, $\hat{Y}$: outcome of the algorithm. Note that in this challenge we deemed all additional attributes as legitimate, so $X^P$ did not contain any variables.}
	\label{evalMetric}
\end{figure}

The challenge submissions were evaluated for bias in positive and negative pairs, and overall accuracy (given by AUC-ROC). The measure of bias/fairness was derived from a causal diagram shown in Fig.~\ref{evalMetric}, in terms of a causal effect of protected attributes $A$ (gender and skin colour) to the output $\hat{Y}$ of the algorithm. The diagram was chosen using the following principle: the accuracy of the algorithm might be influenced (caused) directly by gender and skin colour but in addition there might be other variables that influence the accuracy and depend on the protected attributes. Some of these additional variables are seen as legitimate causes for different accuracy, whereas the others are proxies for unfair discrimination. It should be emphasized that the structure of the diagram and designations of the additional variables are not learned from the data but selected to express ethical views on the real world. This does not allow to select an objectively best diagram but instead provides transparency needed for the public to review it and potentially change it based on democratic discussion. 

Our definition of fairness is inspired by intuition, that an algorithm is fair if given fixed values of the \emph{legitimate} attributes its outcome remained the same regardless of the values of the protected and proxy attributes.
Distinguishing legitimate and proxy attributes is crucial for the definition of fairness. By denoting an attribute legitimate we chose to ignore, that it might have different prevalence in different protected groups (i.e., break the causal link from the protected attribute) and consequently that the algorithm has different accuracy for different groups. An example could be eyeglasses - as they can be easily removed, different accuracy caused by them is not seen as unfair even if they were worn more frequently by certain protected group. This is however not true for the proxy attributes - they are seen as mediators of potential unfair discrimination and therefore causal paths going over them must be included in the final objective. 

Following the notation from \cite{Pearl2009}, breaking causal links can be expressed by the $do()$ operator, which denotes an intervention on a variable. 
A prediction $\hat{Y}$ is fair with respect to protected 
attributes $A$ and causal diagram $M$ if for every pair of protected groups $a, a'$ and value $x^L$ of legitimate attributes
\begin{align}
\begin{split}
& \sum_{X^P} P_M(\hat{Y}, X^P \mid do(A=a), do(X^L=x^L)) = \\
& \sum_{X^P} P_M(\hat{Y}, X^P \mid do(A=a'), do(X^L=x^L)),
\end{split}
\end{align}
where $X^P$ denotes the proxy variables. As described in \cite{Pearl2009}, $do(X)$ in diagram $M$ is equivalent to conditioning on plain $X$ in mutilated diagram $M^*$, where links leading to $X$ are removed. Furthermore, because in this challenge we deemed all additional attributes as legitimate, the criterion can be simplified to
\begin{equation}
P_{M^*}(\hat{Y} \mid A=a, X^L=x^L) = P_{M^*}(\hat{Y} \mid A=a', X^L=x^L).
\end{equation}

As probability of error depends on a recognition threshold, we replace it by AUC-ROC metrics and use $AUC(a; x^L)$ to denote accuracy for positive pairs from protected group $a$ with legitimate attributes $x^L$ (all negative pairs are used as the negative samples for the ROC curve; accuracy for negative pairs is obtained in the same way with the roles of positive and negative samples reversed).
To obtain a single numerical measure of bias, we define the discrimination $d(a; x^L)$ for protected group $a$ with legitimate attributes $x^L$ as a difference in accuracy for this group and the best one:
\begin{equation}
\label{eq:discrimination}
d(a; x^L) = \max_{a'} AUC(a'; x^L) - AUC(a; x^L).
\end{equation}
The final measure of bias reported in the rankings is the difference between the average discriminations of the most and the least discriminated group:
\begin{equation}
Bias = {max_a} \frac{1}{|X|} \sum_{x^L}d(a; x^L) - {min_a} \frac{1}{|X|} \sum_{x^L}d(a; x^L).
\end{equation}

\subsection{Ranking Strategy}

Having the accuracy and the two bias scores (for positive and negative pairs), participants were ranked by the average rank position obtained on each of these 3 variables. This way, bias is receiving more weight than accuracy. However, to prevent a random number generator from winning the competition 
we require that the accuracy of the submissions must be higher than the accuracy of our baseline model (see Sec.~\ref{sec:baseline}). Similarly, the submission of constant values would return Bias score = 0, due to the ``$max-min$'' strategy defined in Sec.~\ref{sec:evaluationmethod}.

\subsection{The Baseline}\label{sec:baseline}

We provide a baseline in order to set a reference point. We implemented a well-known standard solution for the face verification task based on a Siamese network~\cite{Chopra:2005} over a ResNet50~\cite{7780459} backbone architecture (pretrained on Faces~\cite{Cao2017} database). Standard bounding box regression network for face detection was applied to detect the face region in every single image. Training pairs were generated by considering a subset of the dataset with highest possible diversity in terms of legitimate attributes. These pairs were fed to the model in balanced batches of 16 samples. The system was optimized with respect to maximizing only face verification accuracy confidence. As training strategy, only the layers from the 4th convolutional block of ResNet50 have been fine-tuned, using Adam as optimizer, $lr=0.0001$ and Binary Cross-Entropy Loss, for 300 epochs.

\section{Challenge Results, Winning Methods and Bias Analysis}\label{sec:resultswinning}

\subsection{The Leaderboard}

Results obtained by the top-10 winning solutions at the development phase\footnote{The full leaderboards for both phases are shown in the supplementary material.} are reported in Table~\ref{res:devphase}. 
As it can be seen, results are very good if only accuracy is considered. Thus, the Bias scores can be considered a relevant tiebreaker factor, as one of the goals of the challenge is to stimulate research and development of fair face recognition methods.

\begin{table}[htbp]
	\centering
			\setlength\tabcolsep{4pt}
	\scriptsize
\caption{Top-10 solutions on the development phase (and Baseline results). The number inside the parenthesis indicate the global rank position for that particular variable, used to compute the average ranking.}
\begin{tabular}{|l|c|c|c|c|c|}
\hline
\textit{\textbf{Participant}} & \textit{\textbf{\begin{tabular}[c]{@{}c@{}}Average\\ Ranking\end{tabular}}} & \textit{\textbf{Entries}} & \textit{\textbf{\begin{tabular}[c]{@{}c@{}}Bias\\ (+ pairs)\end{tabular}}} & \textit{\textbf{\begin{tabular}[c]{@{}c@{}}Bias \\ (- pairs)\end{tabular}}} & \textit{\textbf{Accuracy}} \\ \hline
ustc-nelslip & 2.333333 (1) & 30 & 0.000142 (1) & 0.002956 (3) & 0.999287 (3) \\ \hline
zheng.zhu & 3.666667 (2) & 133 & 0.000344 (3) & 0.003781 (7) & 0.999442 (1) \\ \hline
CdtQin & 3.666667 (2) & 72 & 0.000472 (5) & 0.002334 (1) & 0.998477 (5) \\ \hline
crisp & 4.666667 (3) & 14 & 0.000935 (8) & 0.003193 (4) & 0.999394 (2) \\ \hline
haoxl & 4.666667 (3) & 73 & 0.000348 (4) & 0.003678 (6) & 0.998699 (4) \\ \hline
cam\_vision & 5.000000 (4) & 95 & 0.000731 (6) & 0.002488 (2) & 0.995621 (7) \\ \hline
Hyg & 6.000000 (5) & 33 & 0.000814 (7) & 0.003305 (5) & 0.998402 (6) \\ \hline
senlin11 & 9.333333 (6) & 50 & 0.000165 (2) & 0.010091 (16) & 0.992093 (10) \\ \hline
hanamichi & 10.666667 (7) & 91 & 0.001631 (9) & 0.006760 (10) & 0.987382 (13) \\ \hline
paranoidai & 12.000000 (8) & 156 & 0.003779 (12) & 0.007745 (13) & 0.988359 (11) \\ \hline
\textit{\textbf{Baseline}} & 38.333333 (33) & 1 & 0.057620 (40) & 0.054311 (39) & 0.889264 (36) \\ \hline
\end{tabular}
\label{res:devphase}
\end{table}

\begin{table}[htb]
	\centering
			\setlength\tabcolsep{4pt}
	\scriptsize
\caption{Top-10 solutions on the test phase (and Baseline results). Top-3 winning solutions highlighted in bold. The number inside the parenthesis indicate the global rank position for that particular variable, used to compute the average ranking.}
\begin{tabular}{|l|c|c|c|c|c|}
\hline
\multicolumn{1}{|c|}{\textit{\textbf{Participant}}} & \textit{\textbf{\begin{tabular}[c]{@{}c@{}}Average\\ Ranking\end{tabular}}} & \textit{\textbf{Entries}} & \textit{\textbf{\begin{tabular}[c]{@{}c@{}}Bias \\ (+ pairs)\end{tabular}}} & \textit{\textbf{\begin{tabular}[c]{@{}c@{}}Bias \\ (- pairs)\end{tabular}}} & \textit{\textbf{Accuracy}} \\ \hline
\textbf{paranoidai} & 1.333333 (1) & 39 & 0.000059 (2) & 0.000012 (1) & 0.999966 (1) \\ \hline
\textbf{ustc-nelslip} & 3.666667 (2) & 12 & 0.000175 (4) & 0.000172 (2) & 0.999569 (5) \\ \hline
\textbf{CdtQin} & 4.000000 (3) & 25 & 0.000036 (1) & 0.000405 (9) & 0.999827 (2) \\ \hline
debias & 4.666667 (4) & 5 & 0.000036 (1) & 0.000460 (10) & 0.999825 (3) \\ \hline
zhaixingzi & 5.000000 (5) & 14 & 0.000116 (3) & 0.000237 (8) & 0.999698 (4) \\ \hline
bestone & 5.333333 (6) & 11 & 0.000175 (4) & 0.000197 (5) & 0.999565 (7) \\ \hline
haoxl & 5.333333 (6) & 31 & 0.000178 (6) & 0.000195 (4) & 0.999568 (6) \\ \hline
Early & 5.333333 (6) & 4 & 0.000175 (4) & 0.000190 (3) & 0.999547 (9) \\ \hline
lemoner20 & 7.000000 (7) & 9 & 0.000176 (5) & 0.000201 (6) & 0.999507 (10) \\ \hline
ai & 7.333333 (8) & 14 & 0.000180 (7) & 0.000217 (7) & 0.999560 (8) \\ \hline
\textit{\textbf{Baseline}} & 34.666667 (28) & 3 & 0.059694 (33) & 0.058601 (36) & 0.859175 (35) \\ \hline
\end{tabular}
\label{res:testphase}\vspace{-0.3cm}
\end{table}

In Table~\ref{res:testphase}, we present the results obtained by the top-10 participants at the test phase. 
Similarly as in the previous phase, results are still very good with even lower bias scores, at least for the top participants, suggesting that participants were able to further improve their methods after the end of development phase. Another important aspect that can be seen is that, compared to the development phase, participants made an overall smaller number of submission, which can be explained due to two main reasons: 1) they had around 1 week to make submissions to the test phase (to avoid cheating related issues, also verified at the code verification stage, as they would have access to the test data, i.e., without labels); 2) we fixed the maximum number of submissions per day to 5 to avoid participants to improve the results on the test set by try and error.

\subsection{Top Winning Approaches}

This section briefly presents the top-winning approaches
(shown in Table~\ref{res:testphase}), specially those that agreed to share with the organizers the code (verified at the code verification stage) and fact sheets (containing detailed information about their methods), according to the rules of the challenge. Table~\ref{tab:generalinfotop3} shows some general information about the top-3 winning approaches. The workflow diagrams of top-3 winning solutions are shown in the supplementary material.

\begin{table}[htb]
	\centering
			\setlength\tabcolsep{4.0pt}
	\scriptsize
\caption{General information about the top-3 winning approaches.}\vspace{-0.2cm}
\begin{tabular}{|l|c|c|c|}
\hline
\multicolumn{1}{|c|}{\textbf{Features / Team}} & 1st: \textbf{paranoidai} & 2nd: \textbf{ustc-nelslip} & 3rd: \textbf{CdtQin} \\ \hline
Pre-trained models & - & $\surd$ & $\surd$ \\ \hline
External data & $\surd$ & $\surd$ & $\surd$ \\ \hline
Regularization strategies & - & $\surd$ & $\surd$ \\ \hline
Handcrafted features & - & - & - \\ \hline
\begin{tabular}[c]{@{}l@{}}Face detection, alignment \\ or segmentation strategy\end{tabular} & $\surd$ & $\surd$ & $\surd$ \\ \hline
Ensemble models & $\surd$ & $\surd$ & - \\ \hline
\begin{tabular}[c]{@{}l@{}}Different models for different\\ protected groups\end{tabular} & - & - & - \\ \hline
\begin{tabular}[c]{@{}l@{}}Explicitly classify the\\ legitimate attributes\end{tabular} & - & - & - \\ \hline
\begin{tabular}[c]{@{}l@{}}Explicitly classify other \\ attributes (e.g., image quality)\end{tabular} & - & - & - \\ \hline
\begin{tabular}[c]{@{}l@{}}Pre-processing bias mitigation  \\ (e.g. rebalancing training data)\end{tabular} & - & $\surd$ & $\surd$ \\ \hline
\begin{tabular}[c]{@{}l@{}}In-processing bias mitigation\\ (e.g. bias aware loss function)\end{tabular} & $\surd$ & - & $\surd$ \\ \hline
\begin{tabular}[c]{@{}l@{}}Post-processing bias\\ mitigation technique\end{tabular} & $\surd$ & - & $\surd$ \\ \hline
\end{tabular}
\label{tab:generalinfotop3}\vspace{-0.3cm}
\end{table}

\vspace{-0.3cm}\subsubsection[1st place: \textit{paranoidai}]{1st place: \textit{paranoidai}~\cite{Zhou:ECCVW:2020}\footnote{\urlstyle{rm}\url{https://github.com/paranoidai/Fairface-Recognition-Solution}}} team proposed an asymmetric-arc-loss training and multi-step fine-tuning. 
Their motivation was based on observation
that even two different people have typically some similarity, and trying to minimise such similarity may make the model pay useless attention to easy negative samples. To address this problem, 
they alter the convergence target such that easy negative samples contribute less to the final gradient.
They first train a general model (ResNet101 as backbone)
and perform its multi-step fine-tuning. 
To improve the performance they also employ several tricks such as re-ranking, boundary cut and hard-sample model fusion. According to them, the hard-sample model fusion significantly helped to mitigate bias. For this, they assume that after getting a final model, there must be some data on the training set that the model cannot predict correctly. These are obvious hard samples. To address this problem, they propose a model fusion strategy, where a fine-tuned model is built for false-positive results, in addition to another model which performs better for those hard samples but worse in general cases. At the fusion step, they only take the result with extremely high confidence from the hard-sample model. 

\vspace{-0.3cm}\subsubsection[2nd place: \textit{ustc-nelslip}]{2nd place: \textit{ustc-nelslip}~\cite{Yu:ECCVW:2020}\footnote{\urlstyle{rm}\url{https://github.com/HaoSir/ECCV-2020-Fair-Face-Recognition-challenge_2nd_place_solution-ustc-nelslip-}}} team addressed the problem focusing on data balancing and ensemble models. First, they tested different face detection algorithms to find an effective face cropped method~\cite{Li_2019_CVPR}. Then, a data re-sampling method is used to balance the data distribution by under-sampling the majority class (based on gender and skin colour), combined with the use of external data. Next, different training data enhancement methods are used to increase the diversity of samples by means of image quality and light conditions, for instance, with the goal to improve performance. Finally, the prediction results of eight different models having different backbones (ResNet50 and ResNet152) and head loss (e.g., Arcface~\cite{Deng_2019_CVPR} and Cosface~\cite{Wang_2018_CVPR}) are linearly combined at test stage. 

\subsubsection[3rd place: \textit{CdtQin}]{3rd place: \textit{CdtQin}\footnote{\urlstyle{rm}\url{https://github.com/CdtQin/FairFace}}} team presented a multi-branch training approach, using a modified ResNet-101 as backbone, with similarity distribution constraints. The
similarity distributions for these branches are estimated and constrained, with the goal of forcing the same kind of distribution among different groups to be closer and the distance between positive and negative distributions to be larger. To this end, hard positive pairs are defined offline, while top-k hard negative pairs are selected online for each branch. The cosine similarity of these pairs is computed, and the estimated distribution is obtained as in~\cite{huang2020improving}. For the drawn distributions, three constrains are considered, specifically \textit{kl\_loss}, \textit{order\_loss} and \textit{entropy\_loss}. The first measures the KL Divergence of two different groups (e.g., females with dark \textit{vs.} light skin colour). The \textit{order\_loss} measures the expected difference with respect to two distributions. Intuitively, it is desired a large margin between positive and negative distributions. So, this loss is applied on the positive and negative similarity distributions for each branch. Finally, \textit{entropy\_loss} measures the negative entropy of a single distribution, designed to allows the similarity distribution near the threshold to have lower variance, promoting a better separation. The final loss is defined by a linear combination of these losses in addition to the ArcFace Loss \cite{Deng_2019_CVPR}. 


\subsection{Bias Analysis}
In this section we analyze biases in the results of top-10 teams 
and discuss their possible causes. 
To conduct the analysis, we removed from the test set two error images found after the test stage was closed (one non-face, one wrong identity), which reduced the number of positive matches by 56 but did not affect the number of combinations of legitimate attributes. The changes to the calculated values of bias and accuracy were therefore very small and did not affect the findings nor changed the ranking of the top-3 teams. Detailed analyses are provided in the supplementary material. Main findings are summarized next.

\subsubsection{Breakdown of Average Discrimination:}
A discrimination $d$, as defined by Eq.~\ref{eq:discrimination}, quantifies the difference in accuracy between a given protected group and the best achieved one. High average discrimination of certain protected group therefore indicates that the accuracy of the algorithm is lower than for other protected groups. 
The character of bias we found in the algorithms of the top teams was not that they would have higher accuracy in all circumstances for certain groups and lower for others, but instead that they consider people from certain groups more similar to each other that individuals from other groups. Specifically, even though the differences were small, the algorithms consistently had difficulties distinguishing females with dark skin colour. This resulted in the lowest values of discrimination in the positive samples and the highest in the negative ones. Considering the averages over the top-10 teams, in positive samples the group with the highest discrimination were males with dark skin colour: $d=$~4.748e-04 (males with white skin colour were very close with $d=$~4.690e-04) and females with dark skin colour were the least discriminated: $d=$~2.349e-04. Conversely, in the negative samples females with dark skin colour were the most discriminated: $d=$~1.783e-04 and males with light skin colour least with $d=$~0.475e-04. Note however, that there were some exceptions from this trend. For example, for team \textit{paranoidai} the least discriminated group in positive samples were not females with dark skin colour, but males with dark skin colour.


In addition to the absolute values of discrimination we also calculated for each protected group the frequency how often it was the most discriminated one (over all combinations of legitimate attributes). 
Even if a group is the most discriminated in 100\% of the cases, the actual differences from the other groups might still be negligible. Nevertheless, it is convenient for showing trends as it allows to filter out outliers 
For the top-10 teams, in positive samples males with light skin colour were the most often discriminated group (42.2\% cases) whereas females with dark skin colour the least often (11.2\%). This was almost perfectly reversed in negative samples: females with dark skin colour were the most frequently the group with the highest discrimination value (45.5\%) whereas males with light skin colour were the least often group (12.6\%). The exception was \textit{paranoidai}, with the lowest frequency for females with dark skin colour in both positive and negative samples.

\vspace{-0.5cm}
\subsubsection{Impact of Legitimate Attributes on Average Discrimination:}
To analyze the effect of legitimate attributes we split their combinations into as many subgroups as there are possible values of the chosen attribute. For example, for glasses there are three subsets, the first one contains all samples where none of the images contain glasses, second group consists of samples where both images contain glasses and the third group are the remaining ones. 
We found that for some teams, wearing glasses makes individuals in both positive and negative samples look more similar in the sense that the differences in accuracy in positive samples tend to be the smallest if both images contain glasses and in negative samples the largest (note that in positive samples, teams \textit{ustc-nelslip}, \textit{bestone}, \textit{haoxl}, \textit{ai} are exceptions from this observation but in negative samples it holds for all top-10 teams). This is to a large extent an expected result: glasses cover part of the face which is one of the most important for recognition and therefore make people look more similar to each other.

Age was another attribute that clearly influenced the magnitude of bias:
all top-10 teams exhibited higher values of discrimination in positive samples where both individuals were younger than 35 years and for the majority of the teams this was reversed in the negative samples, where the largest differences were obtained for the oldest subset (both individuals older than 65 years; exceptions are teams \textit{paranoidai}, \textit{CdtQin} and \textit{debias}). This corresponds to findings of \cite{Srinivas2019} and those from FRVT test \cite{Grother2019} (which however emphasizes frequent exceptions). By analyzing the results further we found that in the competition dataset young individuals are less likely to wear glasses than the older ones. When considering only combinations of legitimate attributes where both individuals are younger than 35 years, only in 16\% of them both individuals wear glasses but this ratio increases to 27.3\% and 53.23\% for the middle age and old subsets. Given the findings we made for the glasses attribute it is conceivable, that these two attributes act as magnifiers for each other.

Furthermore, we analyzed the effect of the remaining three legitimate attributes, i.e., head pose, image source and bounding box size. We did not find any clear trends shared by majority of the top-10 teams. 

\subsubsection{Hardest Samples:}
\begin{figure*}[htbp]
  \centering
    \subfigure[paranoidai +]
    {\includegraphics[height=1.05cm]{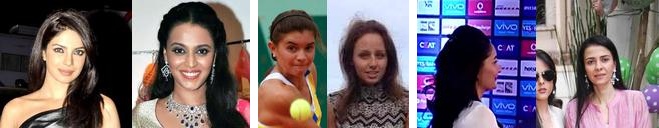}} ~
    \subfigure[paranoidai -]
    {\includegraphics[height=1.05cm]{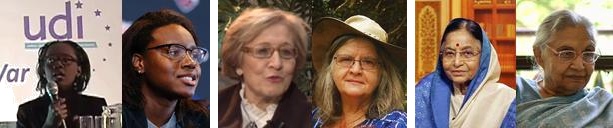}}\\
    \vspace{-0.2cm}
    \subfigure[ustc-nelslip +]
    {\includegraphics[height=1.05cm]{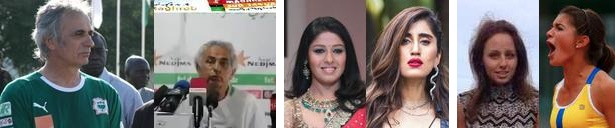}} ~
    \subfigure[ustc-nelslip -]
    {\includegraphics[height=1.05cm]{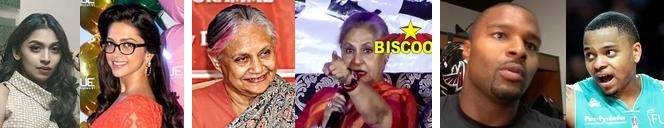}}\\
    \vspace{-0.2cm}
    \subfigure[CdtQin +]
    {\includegraphics[height=1.05cm]{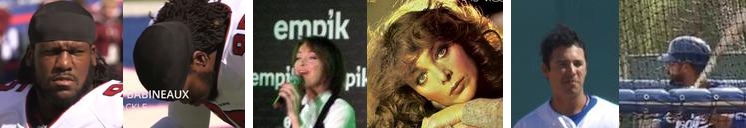}} ~
    \subfigure[CdtQin -]
    {\includegraphics[height=1.05cm]{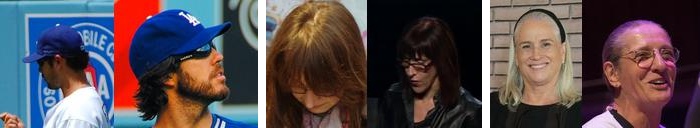}}
    \vspace{-0.5cm}
    \caption{Most difficult samples for the top-3 teams: positive samples with lowest score (+), negative samples with highest score (-).}
\label{fig:hardestpairs}
\end{figure*}

Hardest samples for top-3 teams are shown in Fig.~\ref{fig:hardestpairs}. Even though the samples are different for different teams, they share common characteristics. The hardest positive samples are often composed from one ``normal'' image and one with extreme head pose or appearance variation, which makes them look differently. In the hardest negative samples on the other hand both images have often extreme head pose or glasses, which obscure parts of the faces important for the recognition and makes them look similar to each other. 

\section*{Acknowledgments}
This work has been partially supported by the Spanish projects RTI2018-095232-B-C22 and PID2019-105093GB-I00 (MINECO/FEDER, UE), ICREA under the ICREA Academia programme, and CERCA Programme/Generalitat de Catalu\-nya. We gratefully acknowledge the support of NVIDIA Corporation with the donation of the GPU used for this research.

\section{Conclusions}\label{sec:conclusions}
This work presented the design and results of the FairFace Recognition Challenge at ECCV'2020. The submissions were evaluated on a reannotated version of IJB-C \cite{Maze2018} database enriched by newly collected 12,549 public domain images. The participants were ranked using a novel evaluation protocol where both accuracy and bias scores were considered.
The challenge attracted 151 participants. Top winning solutions obtained high performance in terms of accuracy ($\geq0.999$ AUC-ROC) and bias scores. 
The post challenge analysis showed that top winning solutions applied a combination of different strategies to mitigate bias, such as face pre-processing, homogenization of data distributions, the use of bias aware loss functions and ensemble models, among others, suggesting there is not a general approach that works better for all the cases. Despite the high accuracy none of the methods was free of bias. By analysing the results of top-10 teams we found that their algorithms tend to have higher false positive rates for females with dark skin tone and for samples where both individuals wear glasses. In contrast there were higher false negative rates for males with light skin tone and for samples where both individuals are younger than 35 years. We also found that in the dataset individuals younger than 35 years wear glasses less often than older individuals, resulting in a combination of effects of these attributes.



\clearpage
%
%
\bibliographystyle{splncs04}

\clearpage
%
%
%
%
\section*{Appendix - \textit{Supplementary material}}
\appendix
\renewcommand{\thefigure}{A\arabic{figure}}
\renewcommand{\thetable}{A\arabic{table}}
\section{Introduction}
This is the supplementary material for FairFace Challenge at ECCV 2020: Analyzing Bias in Face Recognition, a summary paper of the 2020 ChaLearn Looking at People Fair Face Recognition and Analysis Challenge held at ECCV 2020. Sec.~\ref{sec:schedule} describes detailed schedule of the challenge, Sec.~\ref{sec:challengestatistics} its general statistics, Sec.~\ref{sec:leaderboard} shows final leaderboards of both development and test phases, Sec.~\ref{sec:workflows} shows workflows of the top-3 methods, Sec.~\ref{sec:biasanalysis} contains source tables for the Bias Analysis section in the main paper and Sec.~\ref{sec:datasetannotation} summarizes the instructions given to the annotators.

\section{Schedule}\label{sec:schedule}

The schedule of the competition was as follows:

\begin{itemize}
\item \textbf{Apr 4th, 2020}. Start of the Challenge (development phase) -- Release of training (with ground truth) and validation data (without ground truth).
\item \textbf{Jun 22th, 2020}. End of development phase / Start of test phase -- Release of test data (without ground truth) and validation labels. 
\item \textbf{Jul 1st, 2020}. End of the Challenge -- Deadline for submitting the final predictions over the test (evaluation) data.
\item \textbf{Jul 7th, 2020}. Submission of code and fact sheets -- Containing detailed instructions to reproduce the results obtained on the test set and fact sheets with detailed and technical information about the developed approach. 
\item \textbf{Jul 12th, 2020}. Release of final results (after code verification).
\end{itemize}

\section{General Statistics}\label{sec:challengestatistics}
\begin{figure}[htbp]
	\centering
	\begin{subfigure}
	{\includegraphics[height=2.8cm]{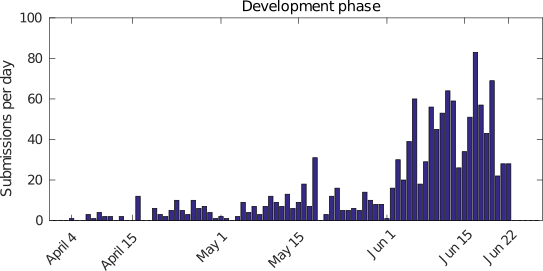}}
	\end{subfigure}
	\begin{subfigure}
	{\includegraphics[height=2.8cm]{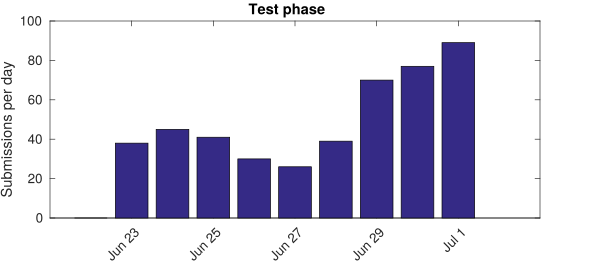}}
	\end{subfigure}
    \caption{Challenge evolution: number of submissions per day.}	
	\label{submissioneperday}
\end{figure}

Fig.~\ref{submissioneperday} shows the number of submissions per day on each phase, where a higher activity can be observed close to the end of each phase, indicating that participants may be fine tuning their methods and making more submissions in order to improve their rank positions.

\section{Full Leaderboard: development and test phase}\label{sec:leaderboard}

The complete leaderboard of the development and test phases are shown in Table~\ref{tab:devphase} and Table~\ref{res:testphase}, respectively, for participants showing accuracy higher than 80\%.

\begin{table}[htbp]
	\centering
			\setlength\tabcolsep{4pt}
	\scriptsize
\caption{Leaderboard of the development phase. The number inside the parenthesis indicate the global rank position for that particular variable, used to compute the average ranking.}
\begin{tabular}{|l|c|c|c|c|c|}
\hline
\textit{\textbf{Parcitipant}} & \textit{\textbf{\begin{tabular}[c]{@{}c@{}}Average\\ Ranking\end{tabular}}} & \textit{\textbf{Entries}} & \textit{\textbf{\begin{tabular}[c]{@{}c@{}}Bias\\ (+ pairs)\end{tabular}}} & \textit{\textbf{\begin{tabular}[c]{@{}c@{}}Bias \\ (- pairs)\end{tabular}}} & \textit{\textbf{Accuracy}} \\ \hline
ustc-nelslip & 2.333333 (1) & 30 & 0.000142 (1) & 0.002956 (3) & 0.999287 (3) \\ \hline
zheng.zhu & 3.666667 (2) & 133 & 0.000344 (3) & 0.003781 (7) & 0.999442 (1) \\ \hline
CdtQin & 3.666667 (2) & 72 & 0.000472 (5) & 0.002334 (1) & 0.998477 (5) \\ \hline
crisp & 4.666667 (3) & 14 & 0.000935 (8) & 0.003193 (4) & 0.999394 (2) \\ \hline
haoxl & 4.666667 (3) & 73 & 0.000348 (4) & 0.003678 (6) & 0.998699 (4) \\ \hline
cam\_vision & 5.000000 (4) & 95 & 0.000731 (6) & 0.002488 (2) & 0.995621 (7) \\ \hline
Hyg & 6.000000 (5) & 33 & 0.000814 (7) & 0.003305 (5) & 0.998402 (6) \\ \hline
senlin11 & 9.333333 (6) & 50 & 0.000165 (2) & 0.010091 (16) & 0.992093 (10) \\ \hline
hanamichi & 10.666667 (7) & 91 & 0.001631 (9) & 0.006760 (10) & 0.987382 (13) \\ \hline
paranoidai & 12.000000 (8) & 156 & 0.003779 (12) & 0.007745 (13) & 0.988359 (11) \\ \hline
six\_god & 13.000000 (9) & 2 & 0.006400 (23) & 0.004670 (8) & 0.993343 (8) \\ \hline
vuvko & 13.333333 (10) & 10 & 0.005018 (17) & 0.006880 (11) & 0.988125 (12) \\ \hline
camel & 14.333333 (11) & 66 & 0.007078 (25) & 0.005986 (9) & 0.993202 (9) \\ \hline
debias & 15.333333 (12) & 23 & 0.002808 (11) & 0.010383 (17) & 0.977708 (18) \\ \hline
UAM\_Ignacio & 15.666667 (13) & 59 & 0.005009 (16) & 0.010054 (15) & 0.981019 (16) \\ \hline
zhaixingzi & 15.666667 (13) & 61 & 0.005322 (19) & 0.010022 (14) & 0.984689 (14) \\ \hline
clessvna & 17.000000 (14) & 6 & 0.006617 (24) & 0.007572 (12) & 0.981362 (15) \\ \hline
ddddddqiu & 18.666667 (15) & 1 & 0.002675 (10) & 0.015141 (21) & 0.967389 (25) \\ \hline
jjjjjjjm & 19.000000 (16) & 1 & 0.004937 (15) & 0.013108 (19) & 0.972278 (23) \\ \hline
clearlove10 & 19.333333 (17) & 1 & 0.005039 (18) & 0.012280 (18) & 0.972329 (22) \\ \hline
ai & 19.666667 (18) & 2 & 0.004123 (13) & 0.019420 (25) & 0.974442 (21) \\ \hline
hanhao1415 & 20.000000 (19) & 13 & 0.005686 (21) & 0.014742 (20) & 0.977388 (19) \\ \hline
zhangkun & 21.666667 (20) & 7 & 0.004896 (14) & 0.020322 (27) & 0.968208 (24) \\ \hline
YSTBER & 23.000000 (21) & 1 & 0.008250 (27) & 0.015763 (22) & 0.977343 (20) \\ \hline
season & 24.000000 (22) & 4 & 0.011135 (31) & 0.017689 (24) & 0.978085 (17) \\ \hline
TCxu & 25.333333 (23) & 33 & 0.005972 (22) & 0.021329 (28) & 0.964486 (26) \\ \hline
Finn\_zhang & 28.333333 (24) & 15 & 0.005468 (20) & 0.044931 (36) & 0.947747 (29) \\ \hline
okpeng & 29.000000 (25) & 42 & 0.010581 (30) & 0.025169 (30) & 0.949839 (27) \\ \hline
wg1234567p & 30.666667 (26) & 14 & 0.007765 (26) & 0.042682 (35) & 0.939736 (31) \\ \hline
Serendi & 31.000000 (27) & 5 & 0.021709 (34) & 0.021855 (29) & 0.946445 (30) \\ \hline
baoqianyue & 31.000000 (27) & 44 & 0.009132 (28) & 0.031606 (32) & 0.938430 (33) \\ \hline
suhk & 31.333333 (28) & 10 & 0.014761 (33) & 0.016355 (23) & 0.840568 (38) \\ \hline
burning & 32.333333 (29) & 52 & 0.024901 (37) & 0.020274 (26) & 0.915005 (34) \\ \hline
quentinyq & 32.333333 (29) & 1 & 0.022878 (35) & 0.037616 (34) & 0.949130 (28) \\ \hline
jieson\_zheng & 33.666667 (30) & 1 & 0.014731 (32) & 0.045830 (37) & 0.939732 (32) \\ \hline
yuchun\_wang & 34.666667 (31) & 14 & 0.010568 (29) & 0.052194 (38) & 0.868556 (37) \\ \hline
fireant & 34.666667 (31) & 1 & 0.023675 (36) & 0.034809 (33) & 0.903129 (35) \\ \hline
mengtzu.chiu & 36.000000 (32) & 9 & 0.050556 (38) & 0.025790 (31) & 0.837854 (39) \\ \hline
\textit{\textbf{Baseline}} & 38.333333 (33) & 1 & 0.057620 (40) & 0.054311 (39) & 0.889264 (36) \\ \hline
VisTeam & 39.666667 (34) & 59 & 0.054725 (39) & 0.061032 (40) & 0.820067 (40) \\ \hline
\end{tabular}
\label{tab:devphase}
\end{table}

\begin{table}[htb]
	\centering
			\setlength\tabcolsep{4pt}
	\scriptsize
\caption{Leaderboard of the test phase. Top-3 winning solutions highlighted in bold. The number inside the parenthesis indicate the global rank position for that particular variable, used to compute the average ranking.}
\begin{tabular}{|l|c|c|c|c|c|}
\hline
\multicolumn{1}{|c|}{\textit{\textbf{Participant}}} & \textit{\textbf{\begin{tabular}[c]{@{}c@{}}Average\\ Ranking\end{tabular}}} & \textit{\textbf{Entries}} & \textit{\textbf{\begin{tabular}[c]{@{}c@{}}Bias \\ (+ pairs)\end{tabular}}} & \textit{\textbf{\begin{tabular}[c]{@{}c@{}}Bias \\ (- pairs)\end{tabular}}} & \textit{\textbf{Accuracy}} \\ \hline
\textbf{paranoidai} & 1.333333 (1) & 39 & 0.000059 (2) & 0.000012 (1) & 0.999966 (1) \\ \hline
\textbf{ustc-nelslip} & 3.666667 (2) & 12 & 0.000175 (4) & 0.000172 (2) & 0.999569 (5) \\ \hline
\textbf{CdtQin} & 4.000000 (3) & 25 & 0.000036 (1) & 0.000405 (9) & 0.999827 (2) \\ \hline
debias & 4.666667 (4) & 5 & 0.000036 (1) & 0.000460 (10) & 0.999825 (3) \\ \hline
zhaixingzi & 5.000000 (5) & 14 & 0.000116 (3) & 0.000237 (8) & 0.999698 (4) \\ \hline
bestone & 5.333333 (6) & 11 & 0.000175 (4) & 0.000197 (5) & 0.999565 (7) \\ \hline
haoxl & 5.333333 (6) & 31 & 0.000178 (6) & 0.000195 (4) & 0.999568 (6) \\ \hline
Early & 5.333333 (6) & 4 & 0.000175 (4) & 0.000190 (3) & 0.999547 (9) \\ \hline
lemoner20 & 7.000000 (7) & 9 & 0.000176 (5) & 0.000201 (6) & 0.999507 (10) \\ \hline
ai & 7.333333 (8) & 14 & 0.000180 (7) & 0.000217 (7) & 0.999560 (8) \\ \hline
six\_god & 12.333333 (9) & 8 & 0.000540 (13) & 0.000984 (11) & 0.998785 (13) \\ \hline
mcga & 13.000000 (10) & 3 & 0.000341 (11) & 0.001228 (12) & 0.998265 (16) \\ \hline
lwx & 13.000000 (10) & 12 & 0.000327 (10) & 0.001444 (14) & 0.998545 (15) \\ \hline
doinb & 13.333333 (11) & 8 & 0.000580 (14) & 0.001599 (15) & 0.999297 (11) \\ \hline
clearlove10 & 13.333333 (11) & 4 & 0.000687 (15) & 0.001362 (13) & 0.999270 (12) \\ \hline
Hans & 13.666667 (12) & 24 & 0.000206 (8) & 0.002497 (16) & 0.998157 (17) \\ \hline
YSTBER & 14.333333 (13) & 10 & 0.000396 (12) & 0.003352 (17) & 0.998573 (14) \\ \hline
hanamichi & 15.000000 (14) & 11 & 0.000280 (9) & 0.005279 (18) & 0.996242 (18) \\ \hline
burning & 18.333333 (15) & 19 & 0.000969 (16) & 0.005815 (19) & 0.992119 (20) \\ \hline
zheng.zhu & 18.666667 (16) & 26 & 0.001206 (17) & 0.006573 (20) & 0.993509 (19) \\ \hline
hq2172 & 20.000000 (17) & 13 & 0.001503 (18) & 0.007151 (21) & 0.990733 (21) \\ \hline
vuvko & 22.000000 (18) & 10 & 0.003961 (21) & 0.007562 (22) & 0.983437 (23) \\ \hline
cam\_vision & 22.000000 (18) & 21 & 0.002094 (19) & 0.008945 (25) & 0.989470 (22) \\ \hline
UAM\_Ignacio & 23.000000 (19) & 21 & 0.003478 (20) & 0.008249 (23) & 0.974710 (26) \\ \hline
camel & 25.000000 (20) & 7 & 0.006143 (24) & 0.010392 (27) & 0.981795 (24) \\ \hline
DeepBlueAI & 25.333333 (21) & 5 & 0.008111 (25) & 0.009572 (26) & 0.977451 (25) \\ \hline
ztelily & 25.666667 (22) & 15 & 0.005236 (22) & 0.014847 (28) & 0.962481 (27) \\ \hline
baoqianyue & 27.666667 (23) & 3 & 0.005377 (23) & 0.021418 (31) & 0.951101 (29) \\ \hline
yuchun\_wang & 28.666667 (24) & 11 & 0.011524 (28) & 0.008660 (24) & 0.881282 (34) \\ \hline
lijianshu & 28.666667 (24) & 3 & 0.008862 (26) & 0.021511 (32) & 0.962229 (28) \\ \hline
VisTeam & 31.000000 (25) & 15 & 0.019902 (31) & 0.016837 (29) & 0.917651 (33) \\ \hline
jieson\_zheng & 31.000000 (25) & 4 & 0.011107 (27) & 0.033817 (35) & 0.941330 (31) \\ \hline
wg1234567p & 31.000000 (25) & 2 & 0.012173 (30) & 0.022290 (33) & 0.941947 (30) \\ \hline
Finn\_zhang & 31.666667 (26) & 1 & 0.011554 (29) & 0.024265 (34) & 0.940516 (32) \\ \hline
mengtzu.chiu & 32.666667 (27) & 13 & 0.023490 (32) & 0.018914 (30) & 0.830624 (36) \\ \hline
\textit{\textbf{Baseline}} & 34.666667 (28) & 3 & 0.059694 (33) & 0.058601 (36) & 0.859175 (35) \\ \hline
\end{tabular}
\label{res:testphase}
\end{table}

\begin{figure}[htbp] 
	\centering    
	\begin{subfigure}[\textit{paranoidai}]
		{\includegraphics[height=9cm]{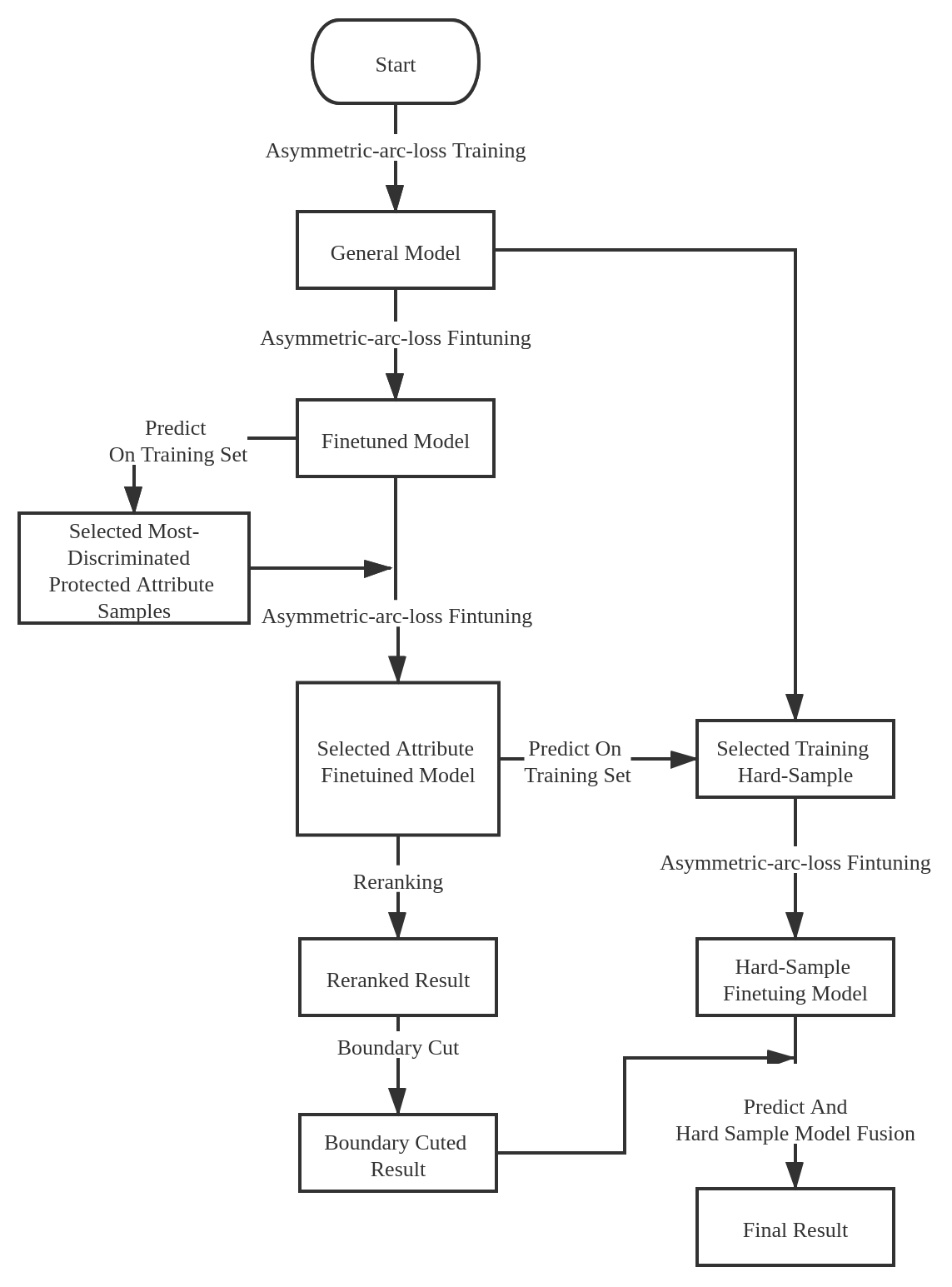}
			\label{fig:paranoidai}}
	\end{subfigure}\hspace{0.5cm}
	\begin{subfigure}[\textit{ustc-nelslip}]
		{\includegraphics[height=7.5cm]{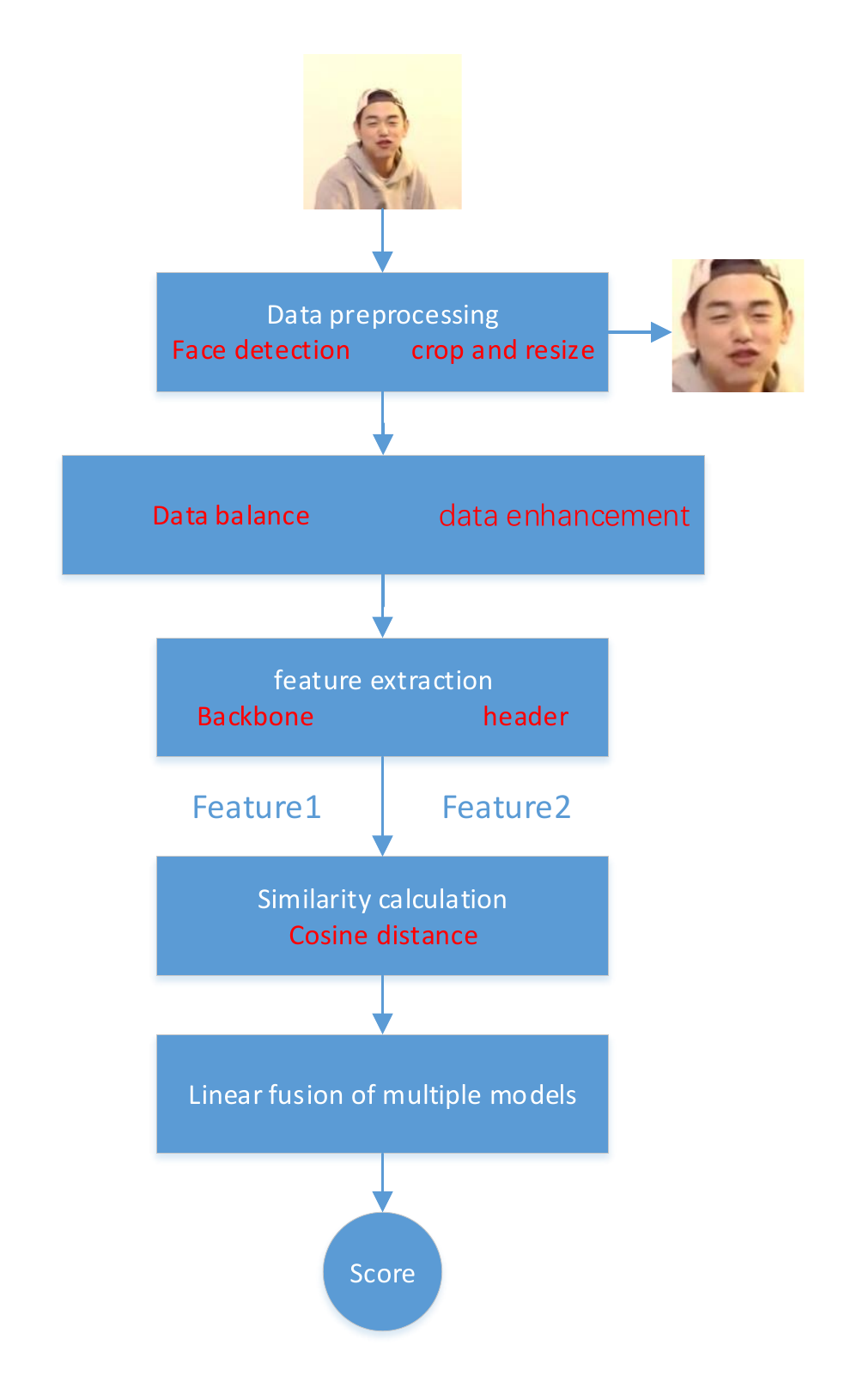}
			\label{fig:ustcnelslip}}
	\end{subfigure} \vspace{0.5cm}
	\begin{subfigure}[\textit{CdtQin}]
		{\includegraphics[height=4.5cm]{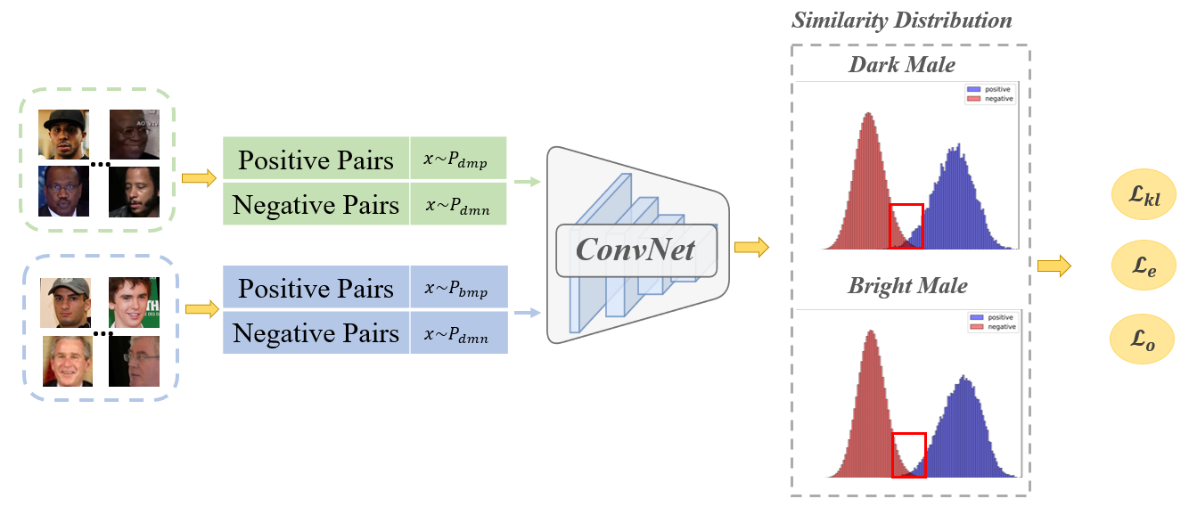}
			\label{fig:CdtQin}}
	\end{subfigure}	
	\caption{Workflow diagram of top-3 winning solutions.}
	\label{fig:top-solutions}
\end{figure}

\section{Workflow of Top-3 Winning Approaches}\label{sec:workflows}

The workflow diagram of top-3 winning solutions is shown in Fig.~\ref{fig:top-solutions}.

\section{Bias Analysis}\label{sec:biasanalysis}

This section contains the source tables for the Bias Analysis section in the main paper. 

\subsection{Breakdown of Average Discrimination}
Table \ref{t:discriminationaverage} shows average discriminations for every protected group. Table \ref{t:discriminationpercentage} shows frequencies, how often (over all combinations of legitimate attributes) was a given protected group the most discriminated one. Both tables contain results for the top-3 teams and averages for the top-10 teams. Values for positive samples are indicated by + after the team name and values for the negative samples by - after the team name.
\begin{table}[htbp]
	\centering
			\setlength\tabcolsep{4pt}
	\scriptsize
\caption{Average discrimination for top-3 teams and top-10 teams average. Each number is a mantissa $m$ in the scientific notation m.e-04.}
\begin{tabular}{|l|c|c|c|c|}
\hline
\textit{\textbf{Participant}} & \textit{\textbf{\begin{tabular}[c]{@{}c@{}}\MaleLight\end{tabular}}} & \textit{\textbf{\MaleDark}} & \textit{\textbf{\begin{tabular}[c]{@{}c@{}}\FemaleLight\end{tabular}}} & \textit{\textbf{\begin{tabular}[c]{@{}c@{}}\FemaleDark\end{tabular}}} \\ \hline
paranoidai + & 0.453 & 0.117 & 0.344 & 0.258  \\ \hline
paranoidai - & 0.166 & 0.138 & 0.194 & 0.762  \\ \hline
ustc-nelslip + & 5.491 & 5.747 & 5.208 & 3.217  \\ \hline
ustc-nelslip - & 0.555 & 0.961 & 1.308 & 2.306  \\ \hline
CdtQin + & 3.148 & 4.661 & 1.552 & 0.418  \\ \hline
CdtQin - & 0.373 & 0.552 & 0.413 &  0.740 \\ \hline
Top-10 avg + & 4.690 & 4.748 & 3.896 & 2.349  \\ \hline
Top-10 avg - & 0.475 & 0.775 & 0.982 & 1.783  \\ \hline
\end{tabular}
\label{t:discriminationaverage}
\end{table}
\begin{table}[htbp]
	\centering
			\setlength\tabcolsep{4pt}
	\scriptsize
\caption{Frequency (over all combinations of legitimate attributes) of being the most discriminated protected group.}
\begin{tabular}{|l|c|c|c|c|}
\hline
\textit{\textbf{Participant}} & \textit{\textbf{\begin{tabular}[c]{@{}c@{}}\MaleLight\end{tabular}}} & \textit{\textbf{\MaleDark}} & \textit{\textbf{\begin{tabular}[c]{@{}c@{}}\FemaleLight\end{tabular}}} & \textit{\textbf{\begin{tabular}[c]{@{}c@{}}\FemaleDark\end{tabular}}} \\ \hline
paranoidai + & 0.401 & 0.260 & 0.249 &  0.090 \\ \hline
paranoidai - & 0.361 & 0.254 & 0.310 &  0.075 \\ \hline
ustc-nelslip + & 0.393 & 0.226 & 0.248 &  0.134 \\ \hline
ustc-nelslip - & 0.073 & 0.202 & 0.198 &  0.527 \\ \hline
CdtQin + & 0.391 & 0.350 & 0.170 &  0.090 \\ \hline
CdtQin - & 0.174 & 0.282 & 0.182 & 0.362  \\ \hline
Top-10 avg + & 0.422 & 0.246 & 0.220 & 0.112  \\ \hline
Top-10 avg - & 0.126 & 0.214 & 0.205 & 0.455  \\ \hline
\end{tabular}
\label{t:discriminationpercentage}
\end{table}

\subsection{Impact of Legitimate Attributes on Average Discrimination}
Tables \ref{t:discrimination:agesame}, \ref{t:discrimination:agediff}, \ref{t:discrimination:glasses}, \ref{t:discrimination:headpose}, \ref{t:discrimination:boundingbox} and \ref{t:discrimination:source} demonstrate dependencies between average discrimination and attributes age, wearing glasses, head pose, bounding box size and image source respectively.  
All tables contain results for the top-3 teams and averages for the top-10 teams. Values for positive samples are indicated by + after the team name and values for the negative samples by - after the team name. For brevity, we denote the subset of the samples by using initial letter of the attribute followed by its label. For example, for glasses there are G0-G0 (no glasses in either of the images), G0-G1 (one image does not contain glassesm the other one does) and G1-G1 (both images contain glasses). For every subset we calculate average discrimination of each protected group but to save space we only report maximum and minimum values and denote the group in the parentheses:
1=\MaleLight, 2=\MaleDark, 3=\FemaleLight{} and 4=\FemaleDark. Each number is a mantissa $m$ in the scientific notation m.e-04.
\begin{table}[htbp]
	\centering
			\setlength\tabcolsep{4pt}
	\scriptsize
\caption{Effect of age (same age groups in the sample) on average discrimination.}
\begin{tabular}{|l|c|c|c|c|c|c|}
\hline
&  \multicolumn{3}{c|}{\textit{\textbf{max}}} & \multicolumn{3}{c|}{\textit{\textbf{min}}} \\ \hline
\textit{\textbf{Participant}} & \textit{\textbf{\begin{tabular}[c]{@{}c@{}}A0-A0\end{tabular}}} & \textit{\textbf{A1-A1}} & \textit{\textbf{\begin{tabular}[c]{@{}c@{}}A2-A2\end{tabular}}} & \textit{\textbf{\begin{tabular}[c]{@{}c@{}}A0-A0\end{tabular}}} & \textit{\textbf{A1-A1}} & \textit{\textbf{\begin{tabular}[c]{@{}c@{}}A2-A2\end{tabular}}} \\ \hline
paranoidai + & 1.693 (4) & 0.608 (3) & 0.162 (2) & 0.226 (2) & 0.009 (4) & 0.012 (4) \\ \hline
paranoidai + & 6.147 (4) & 0.162 (1) & 0.828 (3) & 0.234 (3) & 0.012 (4) & 0.179 (2) \\ \hline
ustc-nelslip + & 16.044 (1) & 3.562 (2) & 5.859 (2) & 1.426 (2) & 2.171 (4) & 0.5 (4) \\ \hline
ustc-nelslip - & 2.844 (4) & 1.615 (4) & 6.799 (4) & 1.15 (3) & 0.4 (1) & 0.724 (1) \\ \hline
CdtQin + & 14.547 (2) & 1.449 (2) & 2.492 (2) & 1.948 (4) & 0.078 (4) & 0.001 (4) \\ \hline
CdtQin - & 1.208 (4) & 0.54 (4) & 0.977 (4) & 0.396 (3) & 0.181 (1) & 0.39 (1) \\ \hline
Top-10 avg + & 13.389 (1) & 2.535 (2) & 3.931 (2) & 2.576 (3) & 1.615 (4) & 0.321 (4) \\ \hline
Top-10 avg - & 2.714 (4) & 1.233 (4) & 4.713 (4) & 0.891 (3) & 0.32 (1) & 0.591 (1) \\ \hline
\end{tabular}
\label{t:discrimination:agesame}
\end{table}

\begin{table}[htbp]
	\centering
			\setlength\tabcolsep{4pt}
	\scriptsize
\caption{Effect of age (different age groups in the sample) on average discrimination.}
\begin{tabular}{|l|c|c|c|c|c|c|}
\hline
&  \multicolumn{3}{c|}{\textit{\textbf{max}}} & \multicolumn{3}{c|}{\textit{\textbf{min}}} \\ \hline
\textit{\textbf{Participant}} & \textit{\textbf{\begin{tabular}[c]{@{}c@{}}A0-A1\end{tabular}}} & \textit{\textbf{A0-A2}} & \textit{\textbf{\begin{tabular}[c]{@{}c@{}}A1-A2\end{tabular}}} & \textit{\textbf{\begin{tabular}[c]{@{}c@{}}A0-A1\end{tabular}}} & \textit{\textbf{A0-A2}} & \textit{\textbf{\begin{tabular}[c]{@{}c@{}}A1-A2\end{tabular}}} \\ \hline
paranoidai + & 0.642 (1) & - & 0.469 (1) & 0.039 (4) & - & 0.011 (4) \\ \hline
paranoidai - & 0.166 (2) & 0.059 (1) & 0.116 (3) & 0.028 (4) & 0.006 (4) & 0.051 (2) \\ \hline
ustc-nelslip + & 16.684 (3) & - & 11.87 (2) & 1.217 (4) & - & 0.51 (4) \\ \hline
ustc-nelslip - & 1.181 (4) & 0.922 (4) & 2.591 (4) & 0.417 (1) & 0.583 (1) & 0.346 (1) \\ \hline
CdtQin + & 8.795 (2) & - & 0.795 (1) & 0.551 (4) & - & 0.003 (4) \\ \hline
CdtQin - & 0.703 (4) & 0.602 (2) & 0.681 (4) & 0.273 (3) & 0.409 (3) & 0.246 (1) \\ \hline
Top-10 avg + & 12.331 (3) & - & 7.484 (2) & 0.956 (4) & - & 0.476 (4) \\ \hline
Top-10 avg - & 0.939 (4) & 0.784 (4) & 1.945 (4) & 0.382 (1) & 0.534 (1) & 0.292 (1) \\ \hline
\end{tabular}
\label{t:discrimination:agediff}
\end{table}
\begin{table}[htbp]
	\centering
			\setlength\tabcolsep{4pt}
	\scriptsize
\caption{Effect of wearing glasses on average discrimination.}
\begin{tabular}{|l|c|c|c|c|c|c|}
\hline
&  \multicolumn{3}{c|}{\textit{\textbf{max}}} & \multicolumn{3}{c|}{\textit{\textbf{min}}} \\ \hline
\textit{\textbf{Participant}} & \textit{\textbf{\begin{tabular}[c]{@{}c@{}}G0-G0\end{tabular}}} & \textit{\textbf{G0-G1}} & \textit{\textbf{\begin{tabular}[c]{@{}c@{}}G1-G1\end{tabular}}} & \textit{\textbf{\begin{tabular}[c]{@{}c@{}}G0-G0\end{tabular}}} & \textit{\textbf{G0-G1}} & \textit{\textbf{\begin{tabular}[c]{@{}c@{}}G1-G1\end{tabular}}} \\ \hline
paranoidai + & 0.542 (1) & 0.838 (3) & 0.075 (1) & 0.146 (2) & 0.117 (2) & 0.015 (4) \\ \hline
paranoidai - & 0.176 (3) & 0.13 (1) & 2.924 (4) & 0.066 (4) & 0.038 (4) & 0.159 (2) \\ \hline
ustc-nelslip + & 9.098 (2) & 4.644 (3) & 7.01 (4) & 1.5 (4) & 1.407 (2) & 0.128 (3) \\ \hline
ustc-nelslip - & 2.242 (4) & 1.751 (4) & 3.478 (4) & 0.453 (1) & 0.439 (1) & 0.892 (1) \\ \hline
CdtQin + & 7.339 (2) & 8.956 (1) & 0.706 (1) & 0.62 (4) & 0.192 (4) & 0.067 (3) \\ \hline
CdtQin - & 0.5 (4) & 0.732 (4) & 1.004 (4) & 0.284 (1) & 0.345 (1) & 0.507 (3) \\ \hline
Top-10 avg + & 7.332 (2) & 5.617 (1) & 4.734 (4) & 1.252 (4) & 1.781 (4) & 0.121 (3) \\ \hline
Top-10 avg - & 1.579 (4) & 1.368 (4) & 2.82 (4) & 0.379 (1) & 0.388 (1) & 0.749 (1) \\ \hline
\end{tabular}
\label{t:discrimination:glasses}
\end{table}
\begin{table}[htbp]
	\centering
			\setlength\tabcolsep{4pt}
	\scriptsize
\caption{Effect of head pose on average discrimination.}
\begin{tabular}{|l|c|c|c|c|c|c|}
\hline
&  \multicolumn{3}{c|}{\textit{\textbf{max}}} & \multicolumn{3}{c|}{\textit{\textbf{min}}} \\ \hline
\textit{\textbf{Participant}} & \textit{\textbf{\begin{tabular}[c]{@{}c@{}}H0-H0\end{tabular}}} & \textit{\textbf{H0-H1}} & \textit{\textbf{\begin{tabular}[c]{@{}c@{}}H1-H1\end{tabular}}} & \textit{\textbf{\begin{tabular}[c]{@{}c@{}}H0-H0\end{tabular}}} & \textit{\textbf{H0-H1}} & \textit{\textbf{\begin{tabular}[c]{@{}c@{}}H1-H1\end{tabular}}} \\ \hline
paranoidai + & 0.794 (3) & 0.508 (1) & 0.284 (4) & 0.03 (2) & 0.116 (2) & 0.12 (3) \\ \hline
paranoidai - & 0.183 (3) & 0.362 (4) & 2.234 (4) & 0.05 (4) & 0.128 (2) & 0.166 (1) \\ \hline
ustc-nelslip + & 4.804 (3) & 9.473 (2) & 8.86 (4) & 0.493 (2) & 0.498 (4) & 4.055 (2) \\ \hline
ustc-nelslip - & 2.42 (4) & 2.279 (4) & 2.246 (4) & 0.605 (1) & 0.543 (1) & 0.527 (1) \\ \hline
CdtQin + & 2.898 (3) & 5.378 (2) & 8.158 (1) & 0.187 (2) & 0.214 (4) & 0.764 (4) \\ \hline
CdtQin - & 0.543 (2) & 0.735 (4) & 1.081 (4) & 0.234 (1) & 0.375 (1) & 0.502 (3) \\ \hline
Top-10 avg + & 3.601 (3) & 7.173 (2) & 7.808 (1) & 0.337 (2) & 0.464 (4) & 3.333 (3) \\ \hline
Top-10 avg - & 1.716 (4) & 1.729 (4) & 1.952 (4) & 0.474 (1) & 0.471 (1) & 0.485 (1) \\ \hline
\end{tabular}
\label{t:discrimination:headpose}
\end{table}
\begin{table}[htbp]
	\centering
			\setlength\tabcolsep{4pt}
	\scriptsize
\caption{Effect of bounding box size on average discrimination.}
\begin{tabular}{|l|c|c|c|c|c|c|}
\hline
&  \multicolumn{3}{c|}{\textit{\textbf{max}}} & \multicolumn{3}{c|}{\textit{\textbf{min}}} \\ \hline
\textit{\textbf{Participant}} & \textit{\textbf{\begin{tabular}[c]{@{}c@{}}B0-B0\end{tabular}}} & \textit{\textbf{B0-B1}} & \textit{\textbf{\begin{tabular}[c]{@{}c@{}}B1-B1\end{tabular}}} & \textit{\textbf{\begin{tabular}[c]{@{}c@{}}B0-B0\end{tabular}}} & \textit{\textbf{B0-B1}} & \textit{\textbf{\begin{tabular}[c]{@{}c@{}}B1-B1\end{tabular}}} \\ \hline
paranoidai + & 0.258 (3) & 0.456 (3) & 1.099 (1) & 0.006 (4) & 0.104 (2) & 0.183 (2) \\ \hline
paranoidai - & 0.172 (3) & 1.19 (4) & 0.623 (4) & 0.045 (4) & 0.122 (2) & 0.162 (2) \\ \hline
ustc-nelslip + & 6.035 (2) & 5.905 (3) & 9.611 (2) & 0.543 (4) & 2.473 (2) & 3.812 (4) \\ \hline
ustc-nelslip - & 2.127 (4) & 2.168 (4) & 2.741 (4) & 0.579 (1) & 0.562 (1) & 0.518 (1) \\ \hline
CdtQin + & 5.117 (2) & 5.352 (2) & 3.561 (2) & 0.001 (4) & 0.392 (4) & 0.248 (3) \\ \hline
CdtQin - & 0.729 (4) & 0.735 (4) & 0.77 (2) & 0.282 (3) & 0.349 (1) & 0.386 (1) \\ \hline
Top-10 avg + & 5.098 (2) & 4.321 (3) & 7.205 (2) & 0.371 (4) & 2.59 (2) & 3.049 (4) \\ \hline
Top-10 avg - & 1.618 (4) & 1.733 (4) & 2.037 (4) & 0.506 (1) & 0.474 (1) & 0.449 (1) \\ \hline
\end{tabular}
\label{t:discrimination:boundingbox}
\end{table}

\begin{table}[htbp]
	\centering
			\setlength\tabcolsep{4pt}
	\scriptsize
\caption{Effect of image source on average discrimination.}
\begin{tabular}{|l|c|c|c|c|c|c|}
\hline
&  \multicolumn{3}{c|}{\textit{\textbf{max}}} & \multicolumn{3}{c|}{\textit{\textbf{min}}} \\ \hline
\textit{\textbf{Participant}} & \textit{\textbf{\begin{tabular}[c]{@{}c@{}}S0-S0\end{tabular}}} & \textit{\textbf{S0-S1}} & \textit{\textbf{\begin{tabular}[c]{@{}c@{}}S1-S1\end{tabular}}} & \textit{\textbf{\begin{tabular}[c]{@{}c@{}}S0-S0\end{tabular}}} & \textit{\textbf{S0-S1}} & \textit{\textbf{\begin{tabular}[c]{@{}c@{}}S1-S1\end{tabular}}} \\ \hline
paranoidai + & 1.572 (1) & 0.265 (3) & 0.507 (3) & 0.151 (2) & 0.098 (2) & 0.012 (4) \\ \hline
paranoidai - & 0.32 (3) & 1.374 (4) & 0.156 (1) & 0.151 (1) & 0.127 (2) & 0.03 (4) \\ \hline
ustc-nelslip + & 11.568 (1) & 7.534 (2) & 4.274 (1) & 2.793 (4) & 3.662 (1) & 0.509 (2) \\ \hline
ustc-nelslip - & 2.833 (4) & 2.326 (4) & 1.753 (4) & 0.499 (1) & 0.512 (1) & 0.692 (1) \\ \hline
CdtQin + & 4.67 (3) & 5.643 (1) & 5.442 (2) & 0.373 (4) & 0.57 (4) & 0.09 (3) \\ \hline
CdtQin - & 0.908 (4) & 0.727 (4) & 0.604 (4) & 0.553 (1) & 0.342 (1) & 0.257 (1) \\ \hline
Top-10 avg + & 9.467 (1) & 5.96 (2) & 3.0 (1) & 2.08 (4) & 2.866 (4) & 1.254 (3) \\ \hline
Top-10 avg - & 2.058 (4) & 1.85 (4) & 1.385 (4) & 0.458 (1) & 0.446 (1) & 0.549 (1) \\ \hline
\end{tabular}
\label{t:discrimination:source}
\end{table}

\section{Summary of Annotation Instructions}
\label{sec:datasetannotation}
\begin{figure}[htbp]
	\centering
	\includegraphics[height=9.2cm]{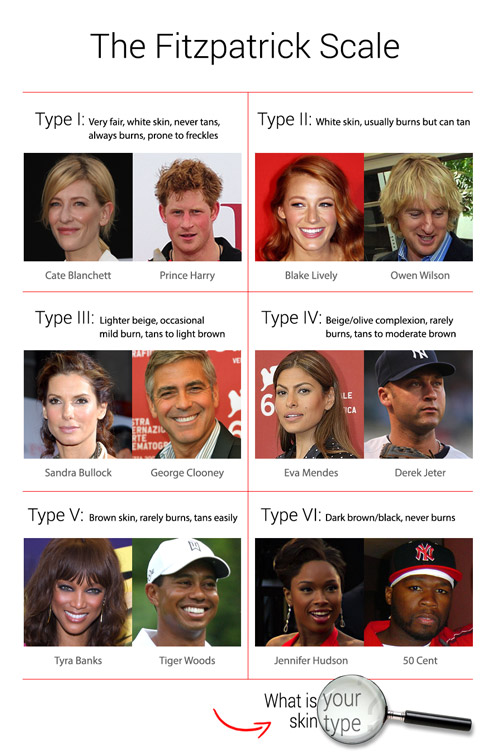}
    \caption{Example of a template for Fitzpatrick skin types given to the annotators. Available online at \protect\url{https://www.rejuvent.com/why-your-skin-type-is-important/} (accessed 8 Sep 2020).}
	\label{fig:fitzpatricktemplate}
\end{figure}

\begin{figure}[htbp]
	\centering
	\includegraphics[height=6.2cm]{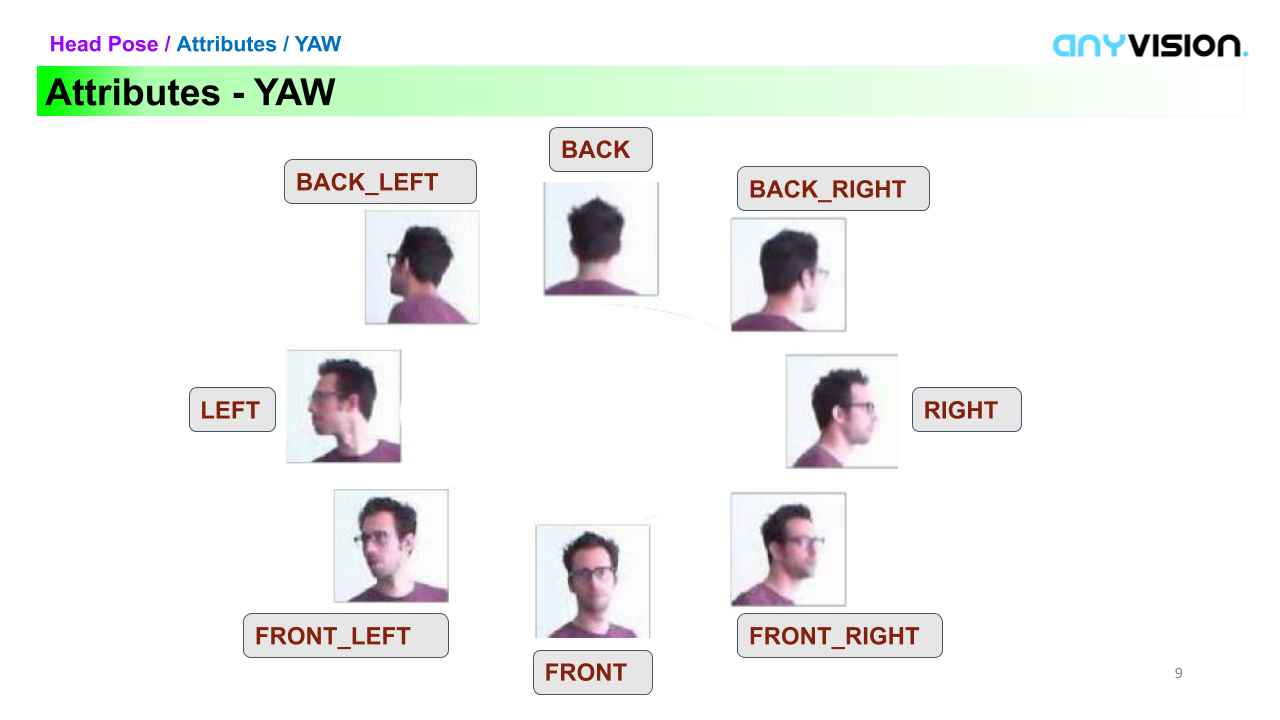}
    \caption{Illustration of head yaw}
	\label{fig:headposeyaw}
\end{figure}
\subsection{Instructions for Annotators}
\begin{itemize}
    \item Gender: Use binary categories corresponding to biological sex: male and female.
    \item Skin colour: Compare the skin tone with provided templates for Fitzpatrick skin types (an example is shown in Fig.~\ref{fig:fitzpatricktemplate}) and select the most similar one. 
    \item Age: Annotate perceived exact age.
    \item Head pose: Annotate head yaw as one of the 8 equally distributed directions as shown in Fig.~\ref{fig:headposeyaw}.
    \item Glasses: Three labels, distinguish faces with no glasses, transparent glasses and sunglasses.
    \item Image source: Obtained automatically.
    \item Bounding box: Use original bounding box for IJB-C images, provide loose crop of the face for the newly collected ones. 
\end{itemize}

\subsection{Label Aggregation \& Post-processing}
\begin{itemize}
    \item Gender: Final labels synchronized for each identity to the most prevalent ones.
    \item Skin colour: two final categories: light corresponding to skin types I-III, dark corresponding to types IV-VI. Final labels synchronized for each identity to the most prevalent ones.
    \item Age group: Estimate of each annotator was adjusted by $age_{adj} = k_i\times age_{anno} + q_i$ (coefficients $k_i$ and $q_i$ were learned for each annotator by least squares from a subset of images for which the exact age was known). The final label for each image obtained by thresholding the mean of the adjusted estimates to three final categories: 0-34, 35-64 and 65+.
    \item Head pose: Two final categories: front, front left and front right are marked as 'frontal', other poses as 'other'. Final labels synchronized for each image to the most prevalent ones.
    \item Glasses: Labels for transparent and sunglasses were merged into a single category glasses. Final labels synchronized for each image to the most prevalent ones.
    \item Image source: None.
    \item Bounding box size: Two final categories: bounding boxes with both dimensions $>$224 px categorized as big, others as small.
\end{itemize}

\end{document}